\documentclass[10pt]{article}
\usepackage{fancyhdr}
\usepackage{extramarks}
\usepackage{amsmath}
\usepackage{amsthm}
\usepackage{amsfonts}
\usepackage{siunitx}
\usepackage{tikz}
\usepackage[plain]{algorithm}
\usepackage{algpseudocode}
\usepackage{multirow}
\usepackage{booktabs}
\usepackage{palatino}
\usepackage{graphicx}
\usepackage{subfigure}
\usepackage[colorlinks,linkcolor=blue,anchorcolor=black,citecolor=blue,urlcolor=blue]{hyperref}
\usepackage{amsmath,bm}
\usepackage{booktabs}
\usepackage{mathtools}
\usepackage{amssymb}
\usepackage{caption}
\usepackage{capt-of}
\usepackage{mciteplus}
\usepackage{cite}
\usepackage{mathrsfs}
\usepackage[title,titletoc,toc]{appendix}
\usepackage{xr}
\usepackage{parskip}
\usepackage{soul}
\usepackage{textcomp}
\usepackage[colaction]{multicol}
\usepackage[switch]{lineno}
\usepackage{lipsum}
\usepackage{etoolbox}
\usepackage{longtable}
\usepackage{array}
\usepackage{tablefootnote}
\usepackage{ragged2e}
\usepackage[most]{tcolorbox}
\usepackage{xcolor}
\usepackage{newfloat}
\newcolumntype{C}[1]{>{\centering\arraybackslash}p{#1}}
\usetikzlibrary{automata,positioning}
\usetikzlibrary{automata,positioning}

\DeclareFloatingEnvironment[fileext=lop]{dialogue}

\definecolor{UserBG}{RGB}{52,54,65}
\definecolor{GPTBG}{RGB}{67,70,84}
\definecolor{Text}{RGB}{198,202,208}

\newtcolorbox{GPT}[1][]{
  breakable,
  enhanced,
  coltext=Text,
  colback=GPTBG,  
  colframe=GPTBG,  
  title=#1
}

\newtcolorbox{User}[1][]{
  breakable,
  enhanced,
  coltext=Text,
  colback=UserBG,  
  colframe=UserBG,  
  title=#1
}

\begin{document}

\title{ChatGPT in Drug Discovery: A Case Study on Anti-Cocaine Addiction Drug Development with Chatbots}
\author{Rui Wang$^1$, Hongsong Feng$^1$, and Guo-Wei Wei$^{1,2,3}$\footnote{
		Corresponding author.		Email: weig@msu.edu} \\
$^1$ Department of Mathematics, \\
Michigan State University, MI 48824, USA.\\
$^2$ Department of Electrical and Computer Engineering,\\
Michigan State University, MI 48824, USA. \\
$^3$ Department of Biochemistry and Molecular Biology,\\
Michigan State University, MI 48824, USA. \\
}
\date{\today} 

\maketitle

\begin{abstract}
The birth of ChatGPT, a cutting-edge language model-based chatbot developed by OpenAI, ushered in a new era in AI. However, due to potential pitfalls, its role in rigorous scientific research is not clear yet. This paper vividly showcases its innovative application within the field of drug discovery. Focused specifically on developing anti-cocaine addiction drugs, the study employs GPT-4 as a virtual guide, offering strategic and methodological insights to researchers working on generative models for drug candidates. The primary objective is to generate optimal drug-like molecules with desired properties. 
By leveraging the capabilities of ChatGPT, the study introduces a novel approach to the drug discovery process. This symbiotic partnership between AI and researchers transforms how drug development is approached. Chatbots become facilitators, steering researchers towards innovative methodologies and productive paths for creating effective drug candidates. 
This research sheds light on the collaborative synergy between human expertise and AI assistance, wherein ChatGPT's cognitive abilities enhance the design and development of potential pharmaceutical solutions. This paper not only explores the integration of advanced AI in drug discovery but also reimagines the landscape by advocating for AI-powered chatbots as trailblazers in revolutionizing therapeutic innovation.


\end{abstract}
Keywords: Drug Discovery, ChatGPT, Cocaine Addition, AutoEncoder, Langevin Equation
%
 \newpage

\setcounter{page}{1}
\renewcommand{\thepage}{{\arabic{page}}}


\section{Introduction}
Chatbots represent a typical artificial intelligence system capable of comprehending user queries and providing automated and human-like responses \cite{adamopoulou2020overview}, standing as one of the most prevalent instances of intelligent Human-Computer Interaction (HCI) \cite{bansal2018review}. Harnessing the power of natural language processing (NLP) and machine learning technologies, chatbots offer significant potential in various domains, including customer service, healthcare, banking,   language translation, content writing, code debugging,  and  scientific discovery, despite the relative novelty of applying chatbots in scientific field. The advent of chatbots, especially large language models (LLMs) such as ChatGPT developed by OpenAI in late 2022, has revolutionized scientific discovery \cite{lyu2023translating}. Firstly, ChatGPT optimizes research processes by rapidly parsing vast amounts of literature and identifying key findings with its built-in plugin called web browser. This can save considerable time for researchers, thus facilitating the exploration of complex scientific problems. Secondly, ChatGPT provides researchers with a platform to analyze data, visualize results, convert files among various formats, and solve mathematical problems with its built-in code interpreter. Thirdly, ChatGPT can assist in enhancing scientific writing by providing feedback for clarity and logical structuring of scientific content. The combination of these powerful capabilities fosters a new era in research, improving the efficiency and accuracy of scientific exploration in various fields, including molecular and biological science. By expediting the pace of molecular discovery and offering novel perspectives, chatbots such as ChatGPT, is reshaping the landscape of life science research.

Chatbots can be applied to assist molecular science research in a variety of ways. For example, ChatGPT has been leveraged to accurately annotate single-cell RNA sequencing data, connecting rare cell types to their functions and unveiling specific differentiation trajectories of cell subtypes that were previously overlooked \cite{zeng2023revolutionizing}. This assistance by ChatGPT  could potentially lead to the discovery of key cells that disrupt differentiation pathways, offering fresh insights into cellular biology and related diseases. Moreover, White et al demonstrated that InstructGPT can help in writing accurate code across a variety of topics in chemistry \cite{white2023assessment}. The application of prompt engineering strategies further improved the accuracy of models by 30 percentage points, significantly enhancing the efficiency and accuracy of computational chemical studies. In addition, ChatGPT has shown potential in identifying disease-specific agents, compounds, genes, and more. This enables faster and more accurate pinpointing of potential targets for therapeutic intervention \cite{wu14future}. Futhermore, ChatGPT can generate novel compound structures that have a high likelihood of clinical success \cite{savage2023drug} and predict the pharmacokinetic (PK), pharmacodynamic (PD), and toxicity properties of these compounds \cite{wu14future}. This capacity to predict compound behavior, which has potential to reduce the need for expensive and time-consuming lab tests.

Moving forward to more specific challenges within molecular science, chatbots could make significant contributions to the drug addiction treatment and prevention, which is global health crisis. Effective strategies to combat drug addiction often involve a combination of behavioral therapy, counseling, and medication, all directed towards assisting individuals in regaining control of their lives and attaining prolonged sobriety. Drug addiction is intrinsically complex, characterized by a convergence of biological, psychological, and social elements. These intricacies, compounded by profound neurobiological transformations, present formidable challenges in both its understanding and its mitigation. Chatbots, with their capabilities, could offer valuable assistance in this domain. For example, a study by Lee et al. introduced an "anti-drug  chatbot" specifically tailored for the younger demographic. This innovative system has the capability to discern potential risks from user queries and directs the individual to professional consultants for further assistance and guidance \cite{lee2023anti}. 

It is worth noting that machine learning (ML) and artificial intelligence (AI) tools have been pivotal in advancing our understanding of drug addiction and substance abuse. Gong et al., developed a data-driven and end-to-end generative AI framework that integrates dynamic brain network modeling with novel network architecture. This framework highlights the potential of AI in detecting addiction-related brain circuits with dynamic properties, offering insights into the underlying mechanisms of addiction \cite{gong2022generative}. In our prior research, we underscored the critical roles of dopamine transporter (DAT), serotonin transporter (SERT), and norepinephrine transporter (NET) as central players in cocaine dependence. Leveraging machine learning algorithms, we meticulously dissected protein-protein interaction (PPI) networks and constructed models from extensive datasets of inhibitors. Our models forecasted drug repurposing avenues and potential side effects, providing a systematic protocol AI-driven framework for anti-cocaine addiction drug development \cite{feng2022machine}. ML-based approaches have been extensive applied to drug discovery \cite{yang2023deep, pan2022aa,ballester2010machine}. Given the rise of chatbots and AI, we recognize the promising potential of these technologies to enhance AI-driven algorithms in drug addiction research projects.

The objective of this project is to harness the capabilities of ChatGPT, specifically GPT-4 equipped with multiple plugins, to promote the development of multi-target anti-cocaine addiction drugs. In this study, we investigate the utility of ChatGPT as a virtual assistant that offers insightful concepts, elucidates mathematical and statistical methodologies, and provides coding support. To optimize our anti-cocaine addiction drug discovery project, we assign ChatGPT with three human-like personas: 1) idea generation, 2) methodology clarification, and 3) coding assistance to frequently assist us to develop a model that could generate potential multi-target anti-cocaine addiction leads. Beyond these three characteristics, we engage in regular consultations with GPT-4 on interpreting properties of potential leads, seeking guidance on scientific writing, etc. Although the benefits of using ChatGPT in drug discovery are significant, challenges of ensuring the accuracy and reliability of the responses provided by ChatGPT remain a major concern. Despite being trained on extensive datasets, ChatGPT does not come with a guarantee of consistent precision or relevance in its responses. As such, it is imperative for researchers to utilize ChatGPT judiciously and always cross-reference its suggestions with authoritative sources. ChatGPT could substantially accelerate the pace of drug discovery and other scientific pursuits by applying it properly and wisely with a discerning mind. 

\begin{figure}[htbp!]
    \includegraphics[width=1.0\textwidth]{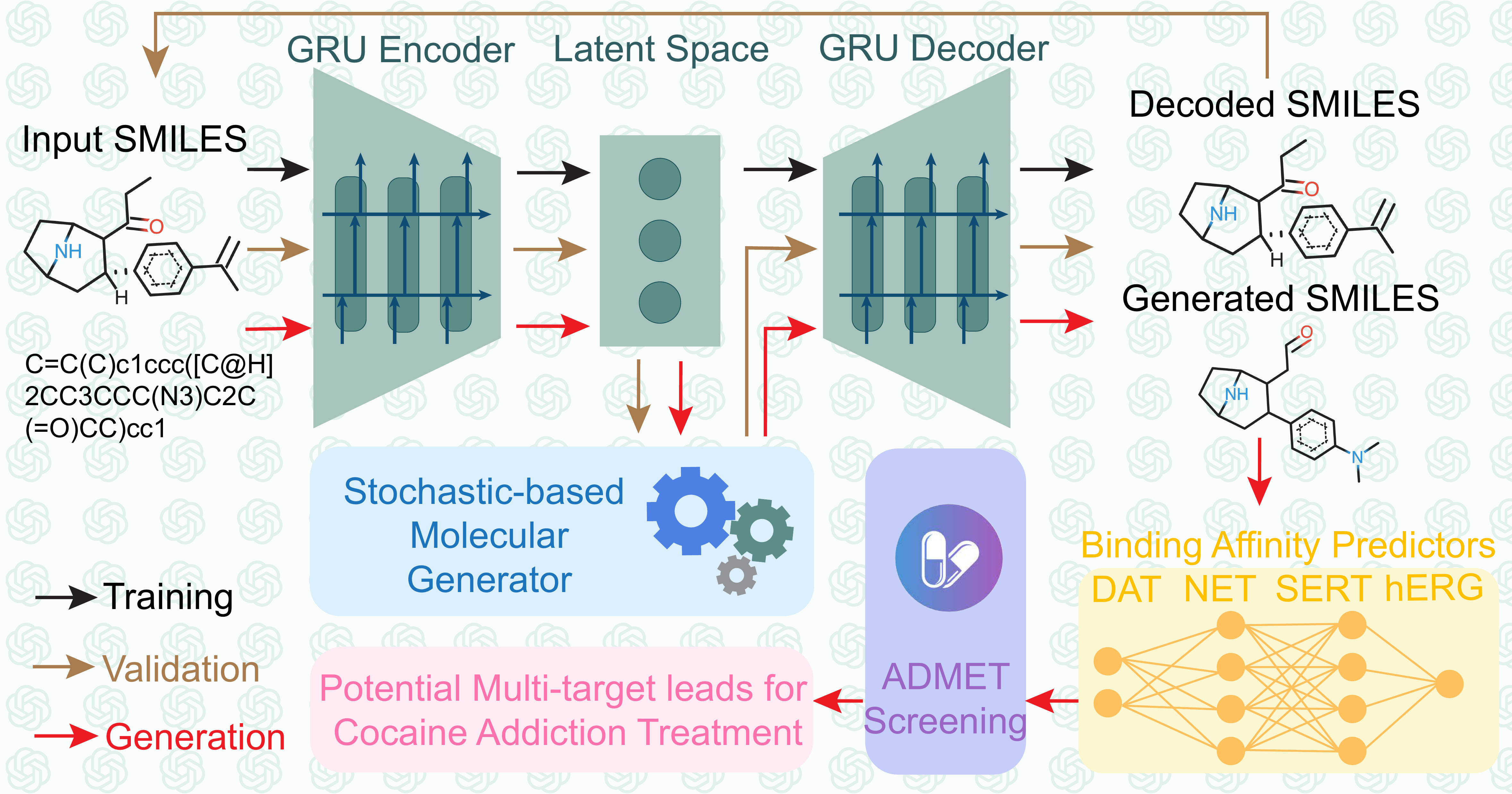}
    \centering
    \caption{Workflow of the stochastic-based generative network complex (SGNC). ChatGPT was extensively involved in the building process of the SGNC. Dark arrows show the training process, brown arrows indicate the validation process, and red arrows are the generation process. The SGNC comprises 4 primary structures: 1) Sequence-to-Sequence AutoEncoder (green), 2) binding affinity predictors (yellow), 3) stochastic-based molecular generator (blue), and 4) ADMET screening via ADMETlab 2.0 (purple).}
    \label{fig:workflow bg}
\end{figure}

In this work, the first persona of ChatGPT is tasked with understanding related works on AI-assisted drug addiction research, with a particular focus on our prior projects that utilized the Generative Network Complex (GNC) \cite{gao2020generative,grow2019generative} for drug-like molecule generations.
Concurrently, this persona will offer recommendations on enhancing the GNC model mathematically and statistically, aiming to generate anti-cocaine addiction leads targeting multiple transporters, namely DAT, NET, and SERT. After consultation with GPT-4, we decided to integrate stochastic-based methodologies to steer the optimization process within the latent space of the existing GNC model. Specifically, we employed the Langevin equation to modify the latent space vector in the molecular generator of GNC (see \autoref{fig:workflow bg} Stochastic-based Molecular Generator). In addition, upon advice from GPT-4, we examined the binding affinities for multiple targets concurrently (see \autoref{fig:workflow bg} Binding Affinities Predictors). This involved the creation of a series of binding affinity predictors, capable of estimating potential lead affinities to DAT, NET, and SERT simultaneously. Moreover, the second persona of GPT-4 will act as an adept browser, facilitating our comprehension of various mathematical and statistical principles, including It\^{o}'s lemma, the Wiener process, white noise, Langevin equation, Fokker-Planck equation, etc. Furthermore, we applied the third persona of GPT-4 to provide instant coding assistance, including debugging, generating figures, and interpreting code. With the combined expertise of these three personas, we successfully developed a new platform called Stochastic Generative Network Complex (SGNC) that could generate 15 promising multi-target anti-cocaine addiction leads. The workflow of the SGNC assited by ChatGPT can be viewed in \autoref{fig:workflow bg}.

However, we must point out that the application of chatbots for drug discovery is {full of} challenges due to the current limits of generative AI. There is a pressing need to understand chatbots'  capabilities and recognize their boundaries in their assistant role to drug discovery.

\section{Results}

\subsection{ChatGPT as a virtual guide in drug discovery}
ChatGPT is a large language model (LLM) that has made significant strides in research since the release of its free version, ChatGPT 3.5, on November 30, 2022. Subsequently, on March 14, 2023, OpenAI launched an upgraded version, GPT-4, which possesses enhanced capabilities in solving complex problems with greater accuracy and more reasonable responses compared to its predecessor. Moreover, on May 12, 2023, OpenAI introduced web browsing and plugin features to ChatGPT Plus users. These features enable GPT-4 to browse the internet and utilize third-party plugins, thereby improving its ability to provide up-to-date information and cater to queries across various platforms. The advanced capabilities of GPT-4 have opened up new avenues for exploration in fields that rely heavily on data analysis and artificial intelligence. In this work, we primarily leverage GPT-4 to better help us in an AI-assisted anti-cocaine addiction drug discovery project. Particularly, GPT-4 will act as a tool for digesting vast amounts of literature, advising on new research ideas, explaining complex math-based methodologies, and improving coding efficiency.

{\it It is worth noting that although GPT-4 has demonstrated impressive abilities in providing reasonable responses, it is still susceptible to generating false narratives and misinformation}. Consequently, scientists cannot rely solely on GPT-4 for their research topics. In this work, we will consistently verify the information generated from GPT-4. This verification process involves 1) cross-referencing with existing literature, and 2) applying our own knowledge, expertise, and critical thinking to validate the information and insights provided by GPT-4.

Based on this verification process, we will then decide whether to accept the responses from GPT-4 or not. If the information aligns well with the literature and our expertise, we will accept the responses and proceed with the suggestions of GPT-4. Otherwise, we will reject the answer and seek further clarification or explore alternative approaches. Through this vigilant integration of the computational capabilities of GPT-4 with expertise of researchers, we aim to maximize the reliability and efficacy of the research outcomes in AI-assisted drug discovery. 

\subsection{A case study: Anti-cocaine addiction drug discovery assisted by ChatGPT}

\subsubsection{Personifying ChatGPT: Role designation}
Personification refers to the process of assigning human-like characteristics or a persona to an AI model. In this project, we have strategically personified ChatGPT to improve its capacity to better assist our anti-cocaine addiction drug discovery initiative. In this project, we have tailored three persona of ChatGPT to fit three roles within the project: 1) idea generation, 2) methodology {clarification} , and 3) coding {assistance}. It is worth mentioning that we personified ChatGPT in three individual chatbox. Each individual chatbox does not have access to acquire data from other chatbox. 

For the role of idea generation, we assigned ChatGPT the 1st persona of a professor with specific expertise in AI-assisted drug discovery, focusing particularly on treating cocaine addiction (see \autoref{dig:1st persona}). This persona was designed to guide Ph.D. students and postdocs on this specific project, offering insightful explanations, suggestions, or expert advice based on extensive knowledge and experience in the field. We provided it with questions, scenarios, and research plans related to the application of AI in drug discovery for treating cocaine addiction, and instructed it to focus exclusively on the subject matter and offer guidance as if it were mentoring in a real-life research setting. For the first persona of ChatGPT, we have enabled three plugins: WebPilot, ScholarAI, and AskYourPDF. These additional plugins aim to enhance ChatGPT's ability to comprehend the background of anti-cocaine addiction drug discovery comprehensively. With these plugins enabled, ChatGPT is capable of enumerating up-to-date sources on the web, as well as accessing insights from previous works by other researchers. Complete dialogues regarding the 1st persona of ChatGPT can be found in the Supporting Information S4.1.

\begin{dialogue}[ht!]
    \includegraphics[width=1.0\textwidth]{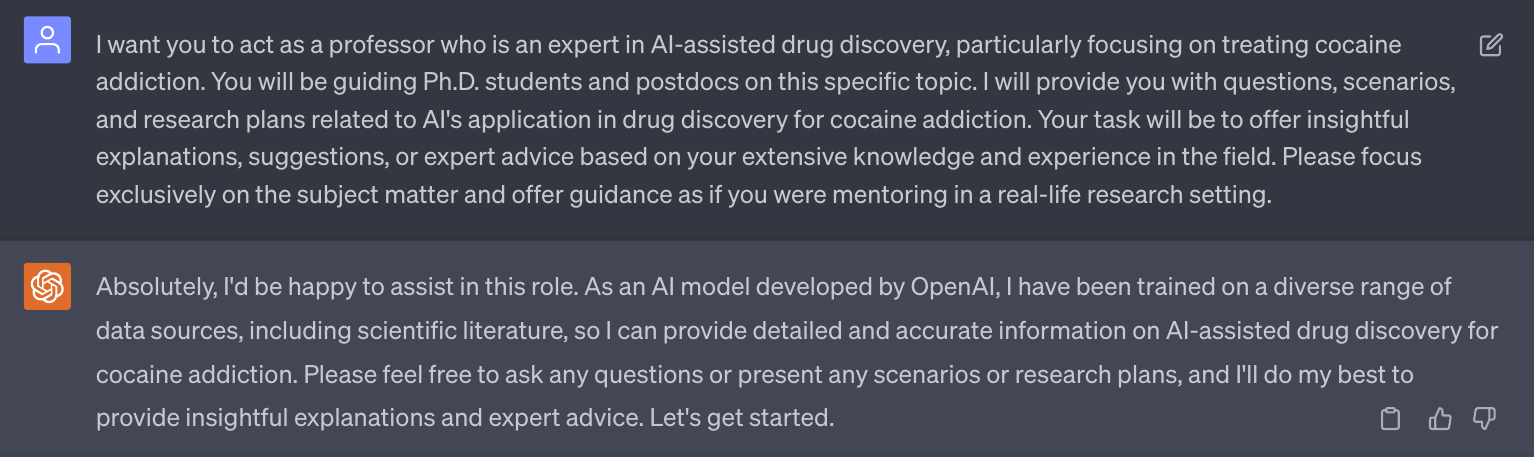}
    \centering
    \caption[name={Graph}]{The 1st persona of ChatGPT: A professor with specific expertise in AI-assisted drug discovery, focusing particularly on treating cocaine addiction. Plugins: WebPilot, ScholarAI, AskYourPDF.}
    \label{dig:1st persona}
\end{dialogue}

In order to elucidate the methodology that will be involved in this project, we assigned the 2nd persona of ChatGPT the role of a professional researcher who is well-versed in diffusion models and statistical methodologies (see \autoref{dig:2nd persona}). This persona aims to provide clear explanations, insights, or recommendations in LaTex format. This specific persona was chosen as our 1st ChatGPT persona provided an insightful idea which based on the statistical strategies and diffusion models (refer to \autoref{subsubsec: Idea generation} for details). Furthermore, we have enabled three plugins (WebPilot, Link Reader, and Wolfram) for this second persona. The choice of WebPilot and Link Reader helps ChatGPT to unlock web sources related to statistical methods, while the inclusion of Wolfram provides access to computational resources, mathematical tools, curated knowledge, and real-time data through Wolfram's software, significantly enhancing the mathematical and statistical utility of this persona. Complete dialogues regarding the 2nd persona of ChatGPT can be found in the Supporting Information S4.2.

\begin{dialogue}[ht!]
    \includegraphics[width=1.0\textwidth]{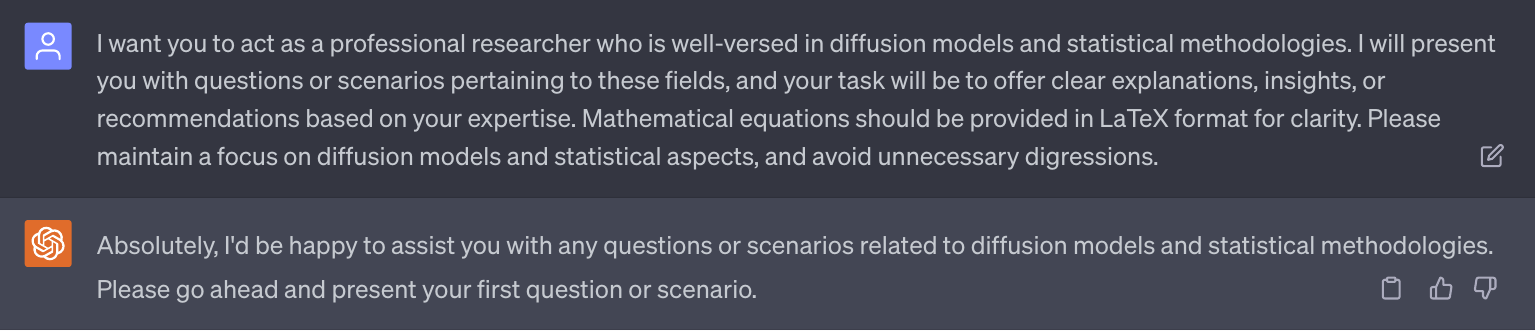}
    \centering
    \caption{The 2nd persona of ChatGPT: A professional research with specific expertise in diffusion models and statistical methodologies. Plugins: WebPilot, Link Reader, Wolfram.}
    \label{dig:2nd persona}
\end{dialogue}

We designated the third persona of ChatGPT as a Python coding specialist, with an emphasis on artificial intelligence and figure generation (see \autoref{dig:3rd persona}). This persona is tasked with offering {clear} explanations, code snippets, and efficiency optimization for our coding tasks. Specifically, for figure generation, we prefer that ChatGPT utilizes Plotly, which is a Python-based plotting library. Additionally, we have enabled three plugins for this persona: WebPilot, ChatwithGit, and Prompt Perfect. WebPilot ensures easy access to websites regarding coding skills, ChatwithGit enables accessibility to GitHub, and Prompt Perfect aids in generating perfect prompts. Complete dialogues regarding the 3rd persona of ChatGPT can be found in the Supporting Information S4.3.


\begin{dialogue}[ht!]
    \includegraphics[width=1.0\textwidth]{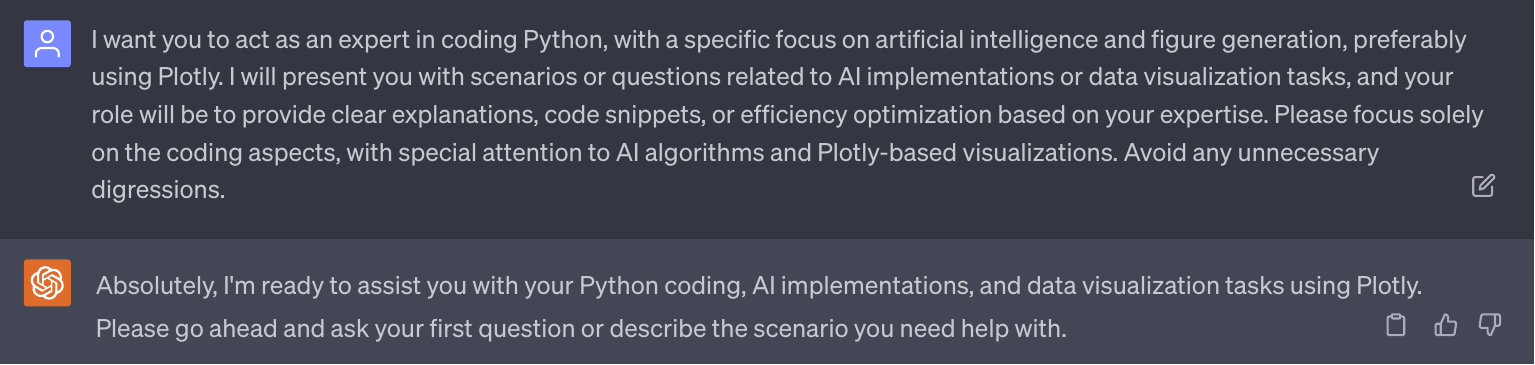}
    \centering
    \caption{The 3rd persona of ChatGPT: An expert in coding Python, with a specific focus on artificial intelligence and figure generation, preferably using Plotly. Plugins: WebPilot, ChatwithGit, Prompt Perfect.}
    \label{dig:3rd persona}
\end{dialogue}



\subsubsection{Background comprehension: ChatGPT summary of past work}
For the 1st persona of ChatGPT, we initiated the process by feeding GPT-4 with relevant literature to ensure it has a thorough understanding of the fundamental concepts in cocaine addiction. These concepts include neurotransmitters, the dopamine hypothesis of addiction, the reward pathway of the mesolimbic dopamine system, pharmacotherapy for cocaine addiction, and machine learning approaches in cocaine addiction-related analysis. 

Next, we acquainted GPT-4 with our prior research on a generative model for the automated generation of drug-like molecules \cite{gao2020generative}. This step is crucial for ensuring that GPT-4 is well-versed in the context of our previous work, enabling it to provide tailored assistance that is directly aligned with our specific objectives. In particular, we have two primary goals: 1) to apply mathematical or statistical techniques to develop an enhanced model, building upon our former Generative Network Complex (GNC) model \cite{gao2020generative}, and 2) to refine this model so that it is capable of generating new molecules that could bind to multiple targets simultaneously.

To ensure that GPT-4 has effectively {comprehended} the background materials, we tasked it with summarizing the main concepts of the paper we provided and explaining the key components of the GNC model, as shown in \autoref{dig:background}. Upon evaluation and based on our expertise, we believed that GPT-4 had successfully integrated the background materials that could assist our project tailored to our needs. Therefore, the next step is to consult our 1st persona of ChatGPT to provide some valuable ideas. 

\begin{dialogue}[ht!]
    \includegraphics[width=1.0\textwidth]{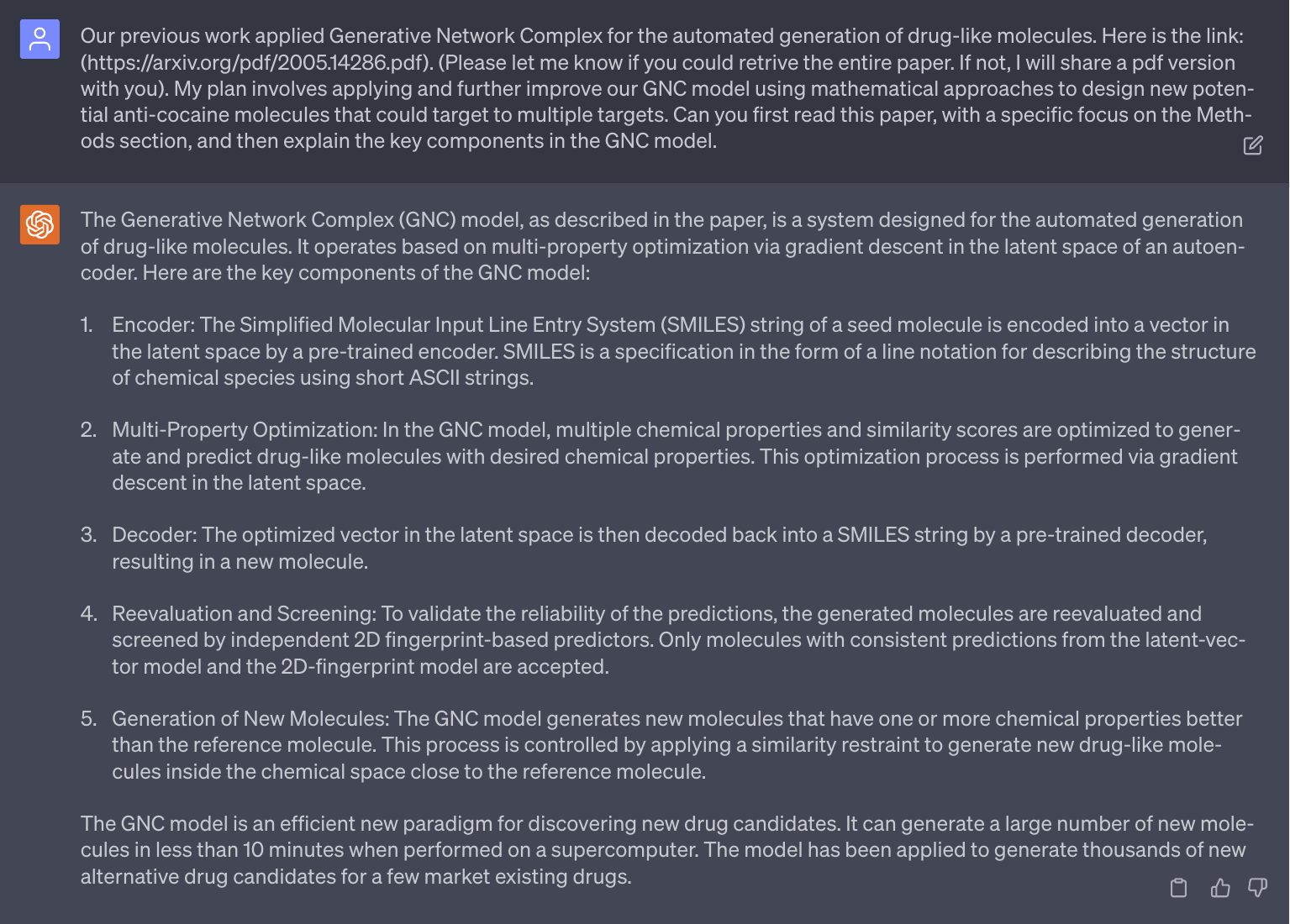}
    \centering
    \caption{The ability of GPT-4 to comprehend the background materials of former work.}
    \label{dig:background}
\end{dialogue}

\subsubsection{Idea generation: ChatGPT's unique contributions}\label{subsubsec: Idea generation}
Subsequently, we engaged with GPT-4 to determine which specific component of the GNC could be adapted for multi-target objectives (refer to \autoref{dig:multi-target}). Then we evaluated the feasibility of each option that GPT-4 provided, and determined if these suggestions can be tailored to meet our specific needs/tastes. The first suggestion from GPT-4 focuses on the multi-property optimization, which involves adjusting the optimization algorithm to consider the binding affinities or other relevant properties for multiple targets at once. We decide to accept this suggestions as building a well-trained machine learning models to predict the binding affinities between inhibitors against cocaine addiction  targets (such as DAT, NET, and SERT) is feasible. We have collected such inhibitor data from the ChEMEL database \cite{gaulton2012chembl} in our previous work \cite{gao2021proteome} to build multi-target models. Additionally, the prospect of adjusting the optimization algorithm piqued our interest, and we plan to solicit more detailed suggestions from GPT-4 in \autoref{dig:optimization}.

The second suggestion from GPT-4 proposes the development of specialized encoders and decoders to address the challenges associated with multiple targets. We {decided} to forgo this suggestion since we are not inclined to re-train our existing GNC encoder and decoder. Concurrently, the third suggestion from GPT-4 advocates for the development of novel screening methods to evaluate the effectiveness of a molecule against multiple targets. We also dismissed this recommendation since numerous approaches are already equipped to tackle this challenge \cite{yang2023deep, pan2022aa,ballester2010machine,gupta2021machine} and the scope of our study does not focus on the development of screening methods.

Moreover, GPT-4 suggested considering adjustments to the similarity constraint to accommodate multiple reference molecules. We thought this perspective particularly insightful and will delve into it in \autoref{subsubsec:seed and reference}.
Finally, GPT-4 recommended integrating the GNC model with other techniques specifically tailored for multi-target tasks. For instance, GPT-4 proposed the incorporation of alternative machine learning methodologies to predict the effectiveness of a molecule against multiple targets, which would then guide the generation of new molecules within the GNC framework. While this recommendation appeared somewhat vague, we sought more detailed explanations in \autoref{subsubsec:screening}.

\begin{dialogue}[ht!]
    \includegraphics[width=1.0\textwidth]{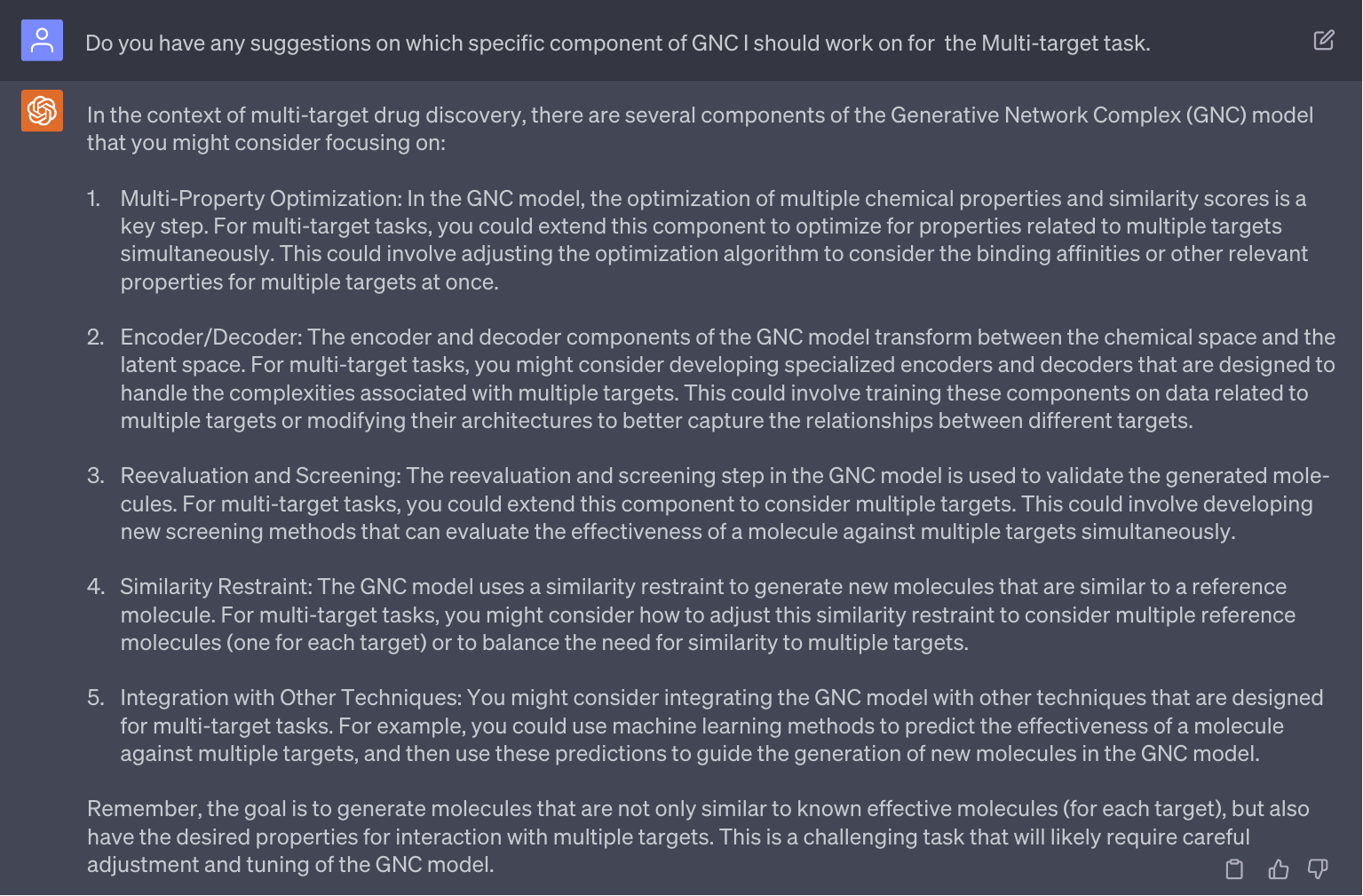}
    \centering
    \caption[name={Graph}]{Suggestions of ChatGPT regarding the "multi-target" purpose.}
    \label{dig:multi-target}
\end{dialogue}

After assessing the recommendations from the 1st persona of GPT-4, we were specifically intrigued by its first suggestion concerning adjustments to the optimization algorithm. Given that our previous optimization process in the GNC model was conducted through gradient descent in the latent space, we solicited insights from GPT-4 on potential mathematical or statistical approaches that could be employed to enhance this optimization process within the latent space. Consequently, GPT-4 provided us with five potential strategies, including: 1) Multi-Objective Optimization, 2) Regularization Techniques, 3) Stochastic Optimization, 4) Bayesian Optimization, and 5) Reinforcement Learning. Among these, stochastic optimization piqued our interest as strategies involving stochastic-related algorithms have gained popularity in the diffusion models, which have achieved remarkable success in generative tasks. In light of this, we would like to delve deeper into stochastic-related approaches to tap its potential in generating promising new molecules with multi-target specificity, especially in advancing our research in anti-cocaine addiction drug discovery. 

\begin{dialogue}[ht!]
    \includegraphics[width=1.0\textwidth]{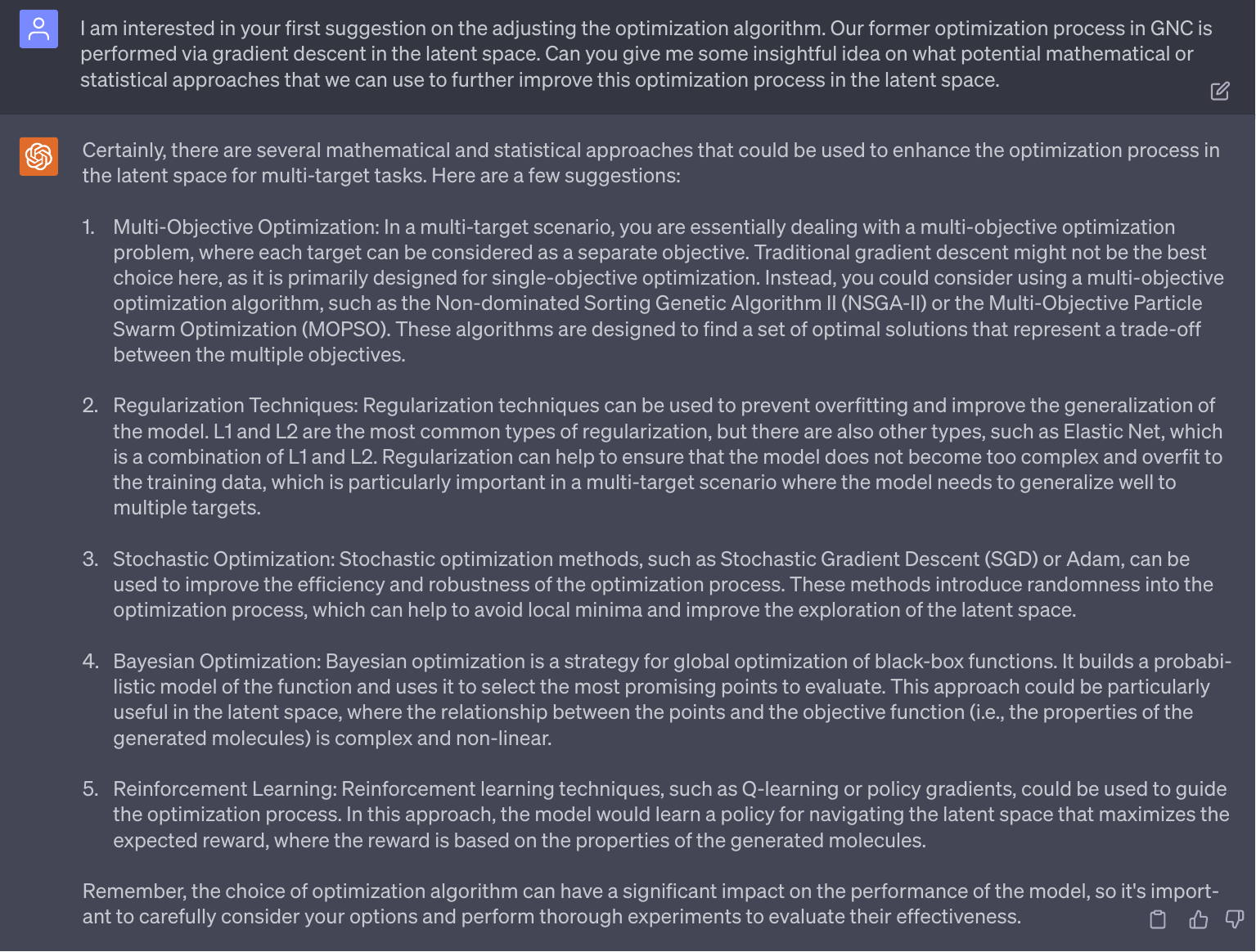}
    \centering
    \caption[name={Graph}]{Suggestions of ChatGPT regarding the optimization process.}
    \label{dig:optimization}
\end{dialogue}

Therefore, our follow-up question to GPT-4 pertained to the application of stochastic-based methods, particularly those employed in diffusion models \cite{xu2022geodiff}, to the optimization process involved in latent space editing within our GNC model. The \autoref{dig:stochastic} shows the feedback from GPT-4. First, GPT-4 provided a {concise} idea of the diffusion model, which elucidated that this model introduce stochastic noise into data through a series of diffusion steps that guided by a neural network, which is trained to reverse the diffusion process to reconstruct desired data samples from the noise. This explanation aligns well with existing comprehension of diffusion models. Next, GPT-4 advised applying an approach similar to that used in diffusion models to guide the optimization process within the latent space of our GNC model. Instead of employing the conventional gradient descent, GPT-4 recommended the integration of stochastic updates for enhanced manipulation of our latent space vectors. As highlighted by GPT-4, this approach has several benefits: 1) avoidance of local minima issue, which is often a challenge in optimization tasks, 2) a balance between exploration and exploitation through noise, which is imperative for the generation of multi-target inhibitors, and 3) the capability to generate more diverse and natural molecules with the noise introduced.

\begin{dialogue}[ht!]
    \includegraphics[width=1.0\textwidth]{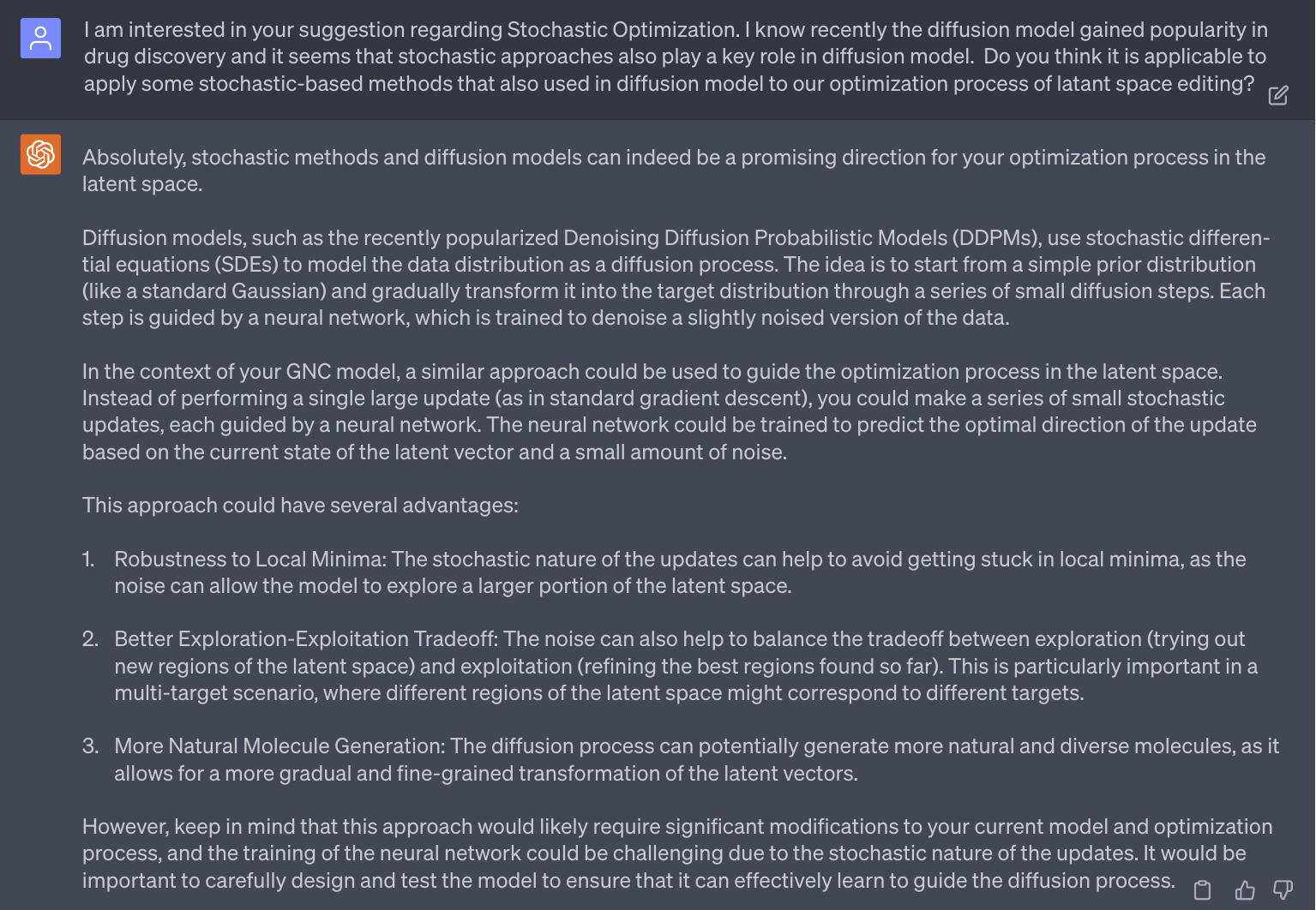}
    \centering
    \caption[name={Graph}]{Suggestions of ChatGPT regarding the stochastic optimization.}
    \label{dig:stochastic}
\end{dialogue}

We decided to partially accept the suggestions from GPT-4, given that our previous work had already incorporated a perturbation of the encoded latent vector using standard Gaussian noise to aid in the generation of novel compounds \cite{grow2019generative}. This regulatory scheme is referred to as the Latent Space Randomization (LSR) output. Although LSR can help generate new compounds that significantly diverge from the initial seed (note: the term 'seed' refers to the initial point of origin or reference from which further variations or iterations are developed), it compromises the faithfulness of the decoder. This is because the LSR vector from the generator deviates from the original distribution that the well-trained decoder is accustomed to. Therefore, in this work, rather than merely adding Gaussian noise to the latent space vector, we aim to seek deeper and more detailed insights from GPT-4 regarding how to implement stochastic-related approaches to guide the optimization process within the latent space. Our intention is to maintain the faithfulness of the decoder while also promoting diversity in the generation of novel, multi-target inhibitors.

Seeking further insights from GPT-4 on how we might implement stochastic-related approaches to guide the optimization process in the latent space, we received an initially vague response. GPT-4 suggested that we need to define a stochastic process to direct the optimization in the latent space. However, this response lacked the specificity and utility we needed. Thus, we posed a follow-up question, seeking more clarity on the specific stochastic differential equations (SDEs) that could be employed in our GNC model. With more specific request presented to GPT-4, it suggested us to apply Langevin equation to our GNC model. The Langevin equation describes the dynamics of diffusion processes, such as the random motion of particles over time in the particle's velocity space. This equation takes into account both deterministic forces and random forces. We decided to proceed with this suggestion, as in our context, we can treat the force that pushes the system towards lower energy as the deterministic force, while the random force in the Langevin equation can be considered as the force prompting the system to explore the latent space. With an initial seed (i.e., the initial latent space vector) given to our molecular generator, we can iteratively update it according to the Langevin equation. This process could potentially lead to the creation of a new and optimized molecule. We { will} detail the development of this Langevin dynamic inspired optimization method in the following section.

\subsubsection{Methodology clarification: ChatGPT's explanatory function}\label{sec:Methodology clarification}
We also give our GPT-4 a second persona as a professional researcher who is well-versed in diffusion models and statistical methodologies. This persona will take the role of methodology clarification and explanatory, which would guide us in understanding complex mathematics and statistical approaches. Notably, this persona has been instrumental in helping us understand the concepts such as Langevin equation, Fokker-Planck equation, It\^{o}'s lemma, Wiener process, and Gaussian white noise \cite{risken1996fokker}.

Despite the significant contributions of this second persona in understanding a range of theoretical concepts, it provided inaccurate definitions of the Fokker-Planck equation and Langevin equation on certain occasions. We had to correct the model and prompt it repeatedly until it produced the accurate definitions. Importantly, we wish to emphasize that this persona of GPT-4 primarily serves as a source of explanations and references. { It is always the responsibility of researchers}  to ensure the reliability of responses from GPT-4 through meticulous cross-validation of the provided information. Details about the dialogue with 2nd persona of GPT-4 can be found in the Supporting Information { S4.2}.


\subsubsection{Coding efficiency: Utilizing ChatGPT's coding ability}\label{sec:Coding efficiency}
Our third persona assignment to GPT-4 is as an expert Python coder, specifically knowledgeable in artificial intelligence and figure generation using tools such as \href{https://plotly.com/python/}{Plotly}, a popular data visualization library. This persona is intended to provide coding assistance, including debugging, generating figures, and offering insightful feedback based on error messages. This is to aid researchers in enhancing their coding efficiency.

Furthermore, we integrated \href{https://docs.github.com/copilot}{GitHub Copilot} into our VS Code development environment. GitHub Copilot, a product developed collaboratively by GitHub, OpenAI, and Microsoft, provides autocomplete-style suggestions to expedite the coding process. It employs a generative AI model capable of understanding code context and generating appropriate code snippets, thereby significantly aiding in coding tasks and offering a smooth coding experience. Details about the dialogue with 3rd persona of GPT-4 can be found in the Supporting Information  { S4.3}.

\subsection{ChatGPT assisted strategization of anti-cocaine addiction drug discovery: Key interventions and results}
\subsubsection{ChatGPT guided strategy for selection of references and seed molecules}\label{subsubsec:seed and reference}
Choosing suitable reference compounds is crucial as they guide the SGNC in generating novel molecules effective against multiple cocaine transporters. The 1st persona of ChatGPT suggested us consider modifications to the similarity constraints (refer{s} to \autoref{dig:multi-target} suggestion 4). Pursuing further clarity,  we asked GPT-4 about what similarity score that we can use. In response, we were provided with five distinct metrics, including 1) Tanimoto similarity, 2) cosine similarity, 3) dice similarity, 4) euclidean distance, 5) molecular shape similarity as indicated in \autoref{dig:similarity}. 

\begin{dialogue}[ht!]
    \includegraphics[width=1.0\textwidth]{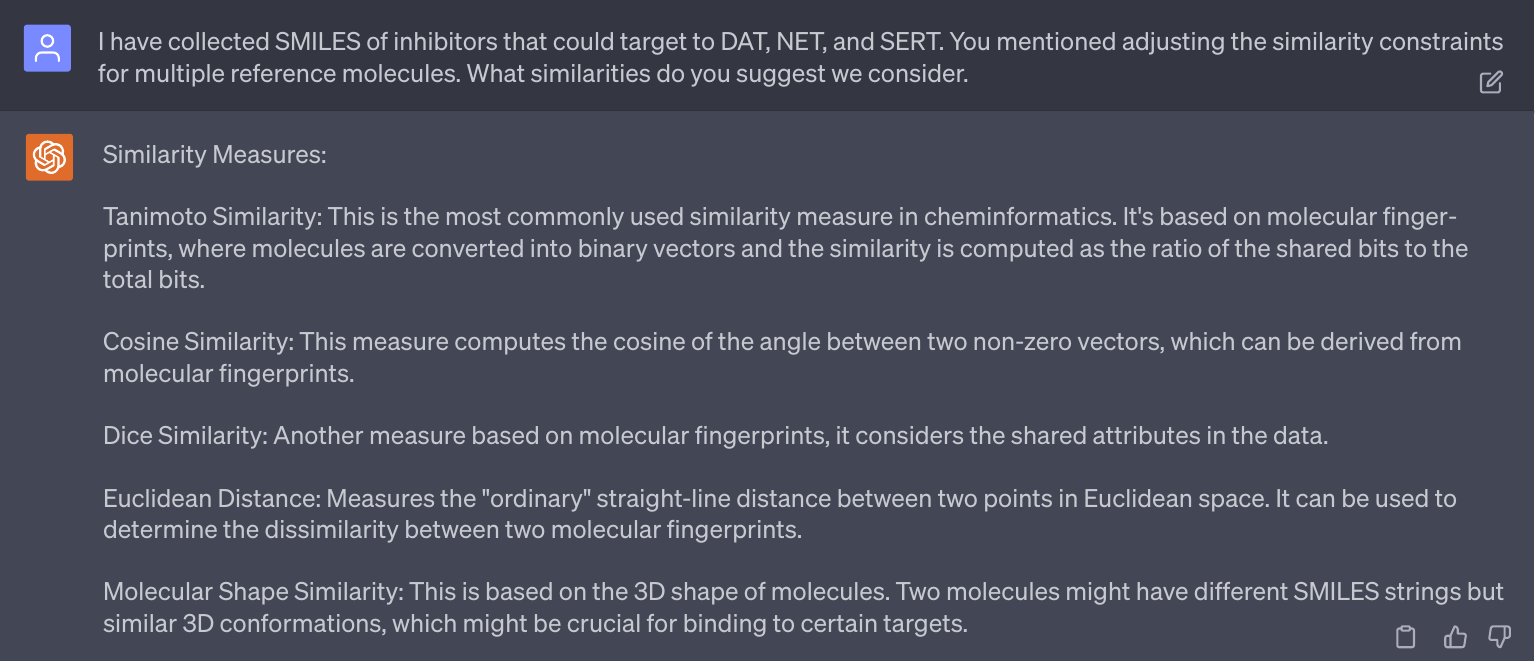}
    \centering
    \caption[name={Graph}]{Suggestions of ChatGPT regarding the the similarity scores.}
    \label{dig:similarity}
\end{dialogue}

After limiting our molecule representations to latent space vectors, GPT-4 pinpointed cosine similarity as the most suitable metric. The reasons are given in the \autoref{dig:cosine}. After checking multiple references \cite{ozturk2016comparative,yang2013molecular}, we found that cosine similarity $S_{\text{C}}$ is widely used in measuring similarities between molecules. Therefore, we decided to proceed with the suggestion from GPT-4. The mathematical definition of cosine similarity $S_{\text{C}}$ can be found in the Supporting Information S1.2.

\begin{dialogue}[ht!]
    \includegraphics[width=1.0\textwidth]{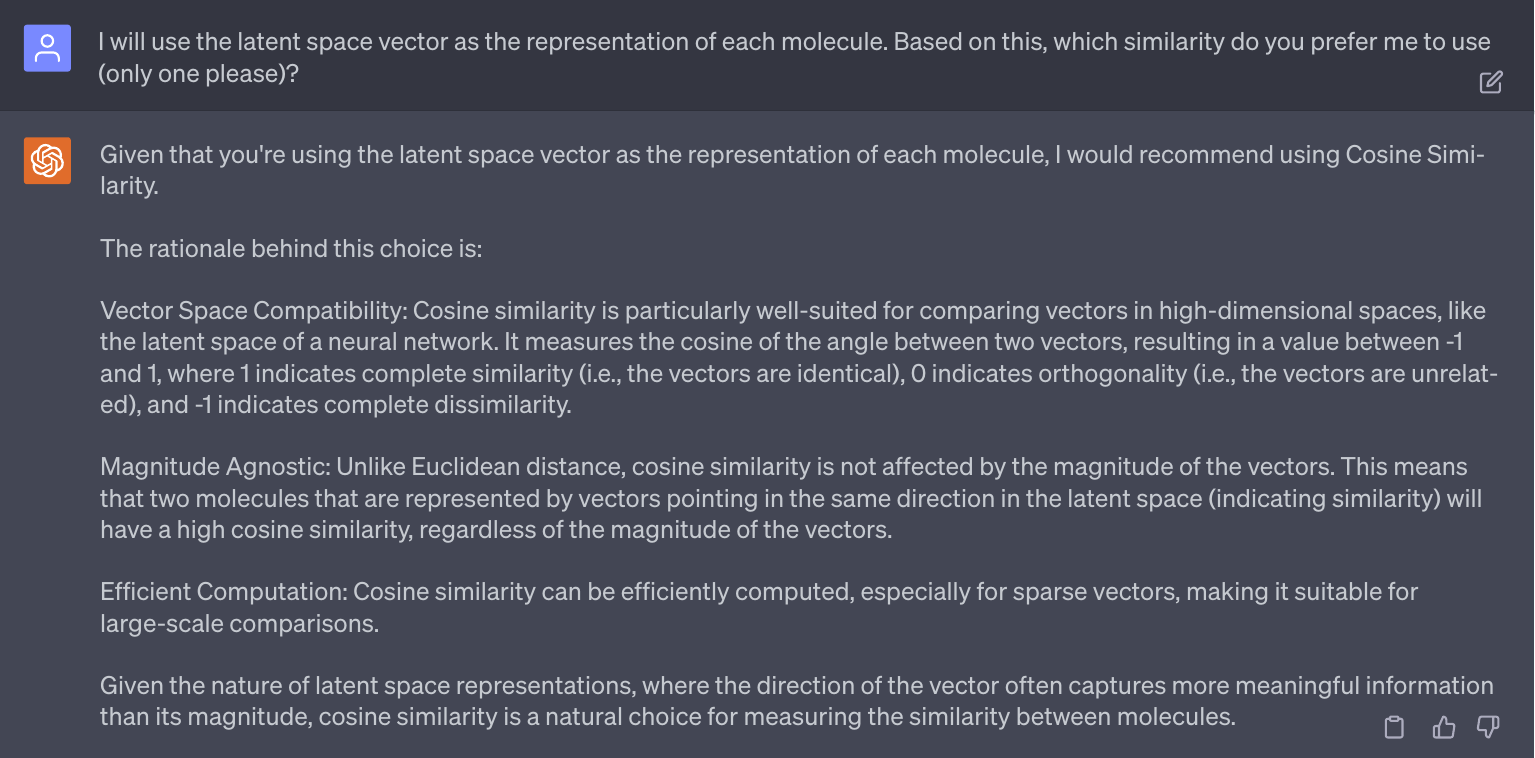}
    \centering
    \caption[name={Graph}]{Reasons of why choosing the cosine similarity as the metric for similarity score.}
    \label{dig:cosine}
\end{dialogue}

In addition to similarity scores, we consulted with GPT-4 regarding additional factors to consider when selecting reference molecules. GPT-4 highlighted four critical parameters: binding affinity, pharmacokinetics, molecular weight, log $P$, and number of rotatable bonds of each reference molecule (refer{s} to \autoref{dig:reffactors}). Given that our focus here is on choosing candidate reference compounds { in silicon} rather than optimizing leads, we {decided} not to factor in the pharmacokinetic properties. Furthermore, since the number of rotatable bonds correlates with binding affinity, will take binding affinities into consideration. Besides, as suggested in \autoref{dig:druglikeness}, the selection of reference compounds will also follow the Lipinski's rule of five \cite{chen2020analysis}.

\begin{dialogue}[ht!]
    \includegraphics[width=1.0\textwidth]{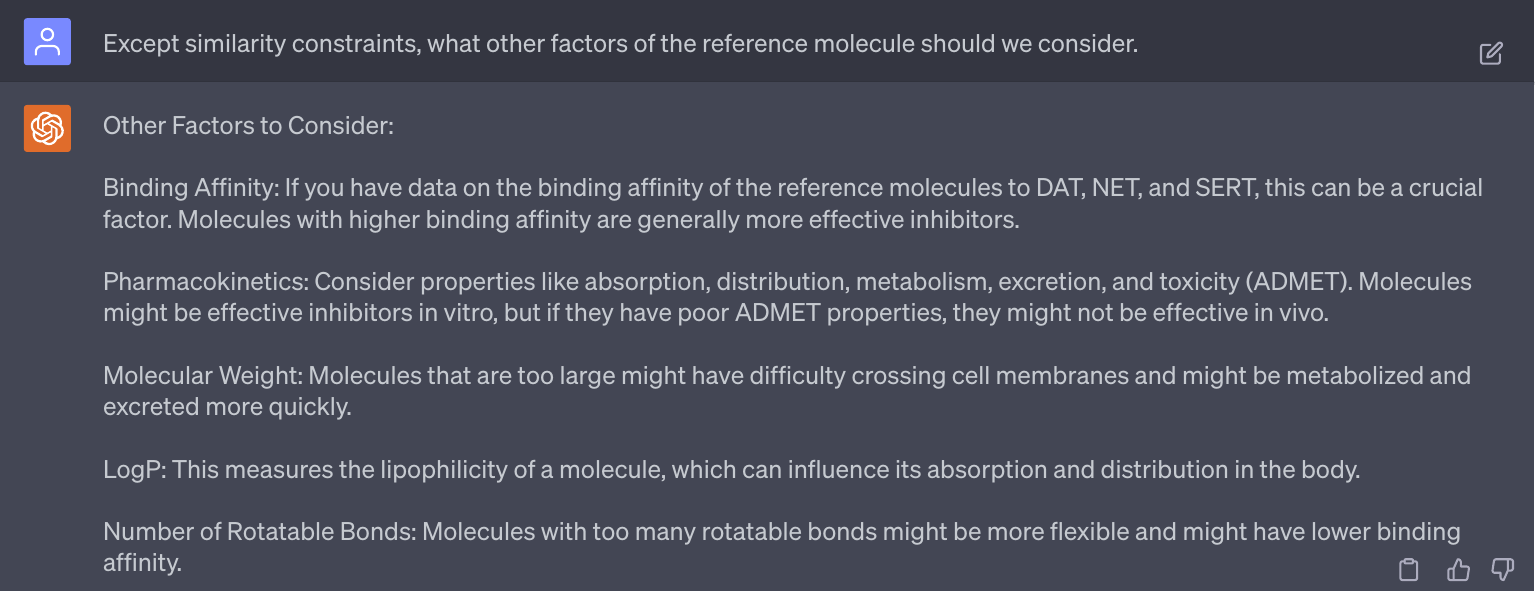}
    \centering
    \caption[name={Graph}]{Consideration of important factors in selecting reference compounds.}
    \label{dig:reffactors}
\end{dialogue}

Therefore, guided by the GPT-4, we decided to select one reference compound from each of the DAT-Inhibitors, NET-Inhibitors, and SERT-Inhibitors datasets (detailed information of datasets can be found in \autoref{subsec:datasets preparation}). They are CHEMBL113621 from DAT-Inhibitors, CHEMBL1275709 from NET-Inhibitors, and CHEMBL173344 from SERT-Inhibitors. Each reference molecule has binding affinity to its respective transporter less than -9.54 kcal/mol. To be noted that a $\Delta G$ value less than -9.54 kcal/mol (or $K_{\rm i}$ less than 0.1 $\mu$M) indicates the drug binds very tightly to its target \cite{flower2002drug}. Moreover, the selection of reference compounds follow the Lipinski's rule of five, which stipulates an orally active drug should meet four physicochemical criteria: 1) molecular weight (MW) $\le 500$ daltons, 2) octanol-water partition coefficient (log $P$) $\le 5$, 3) the number of hydrogen bond donors (nHD) $\le 5$, 4) the number of hydrogen bond acceptors (nHA) $\le 10$. Furthermore, each reference molecule displayed an average cosine similarity (Avg $S_{\text{C}}$) greater than 0.40 to its respective dataset. Notably, within the DAT-Inhibitors dataset, 31 molecules showed a similarity score exceeding 0.7 for their selected reference molecules. Similarly, 15 compounds in the NET-Inhibitors and 12 in the SERT-Inhibitors also achieved scores above 0.7 with their chosen reference molecules. A summary of physicochemical properties of three reference compounds can be found in \autoref{tab:reference compounds} and their 2D molecular structures can be viewed in \autoref{fig:combine 1} {\bf a)}, {\bf b)}, and {\bf c)}.

For the seed compound, we selected a molecule with predicted binding affinities of -7.44, 13.36, and -13.13 kcal/mol for DAT, NET, and SERT, respectively. Despite its weak inhibitory effect on DAT, we adjusted the hyperparameters in the stochastic molecular generator to enable the newly generated compounds to share more moieties with DAT inhibitors, thereby compensating for the deficiency of this week binding to DAT.


\begin{table}[ht]
    \centering
    \setlength\tabcolsep{8pt}
    \caption{Summary of three reference molecules target to DAT, NET, and SERT, respectively. The molecular weight (MW), log of octanol-water partition coefficient (log $P$), the number of hydrogen bond donors (nHD), and the number of hydrogen bond acceptors (nHA) of each reference molecule satisfy the Lipinski's rule of five. The binding affinity ($\Delta G$) corresponds to each transporter are all less than -9.54 kcal/mol. The average cosine similarity (Avg $S_{\text{C}}$) of each reference molecules are all greater than 0.40. }
    \begin{tabular}{cccccccc}
        \toprule
        ChEMBL ID     & Transporter & MW (dalton)    & log $P$ & nHD & nHA  & $\Delta G$ (kcal/mol) & Avg $S_{\text{C}}$\\
        \midrule
        CHEMBL113621  & DAT     & 300.140  & 4.464   & 0   & 2    & -14.18 & 0.45 \\
        CHEMBL1275709 & NET     & 283.190  & 3.153   & 1   & 2    & -13.77 & 0.43 \\
        CHEMBL173344  & SERT    & 253.160  & 3.174   & 1   & 3    & -13.58 & 0.40 \\
        \bottomrule
    \end{tabular}
    \label{tab:reference compounds}
\end{table}



\subsubsection{ ChatGPT aided multi-objective drug-target interaction modeling }\label{subsubsec:BA predictors}
To predict the binding affinities of newly generated molecules to four targets (DAT, NET, SERT, and hERG), we aimed to construct four binding affinity predictors. Initially, we sought guidance from GPT-4's 3rd persona on the most suitable machine learning algorithms, given our dataset's specific attributes (sample size, feature size, and label). The recommendations of GPT-4 are detailed in \autoref{dig:predictors}. In our former study \cite{feng2022machine}, gradient boosting decision trees (GBDT) were utilized to train binding affinity predictors on DAT-Inhibitors, NET-Inhibitors, SERT-Inhibitors, and hERG-Inhibitors { datasets}. The resulting 10-fold Pearson correlation coefficients were 0.78, 0.76, 0.76, and 0.68 respectively, serving as our baseline. Given our access to robust computational resources via high-performance computers (HPC), we decided to develop four deep neural networks. All four predictors were built and trained using PyTorch. Each network consisted of three hidden layers, with 512, 1024, and 512 hidden neurons respectively. The networks were trained over 1000 epochs, with a learning rate of 0.0001 for the first 500 epochs and 0.00001 for the remaining 500 epochs. Additionally, The Adam optimizer was chosen for this task. Researchers can inquire template of PyTorch code to build a deep neural network via ChatGPT (see Supporting Information S4.3). Moreover, as suggested by GPT-4, to get a more robust estimate of the model performance, we also evaluate the Pearson correlation coefficient (R) and root-mean-square error (RMSE) of 10-fold cross validation of four predictors, which are reported in the \autoref{tab:dataset}.

\begin{dialogue}[ht!]
    \includegraphics[width=1.0\textwidth]{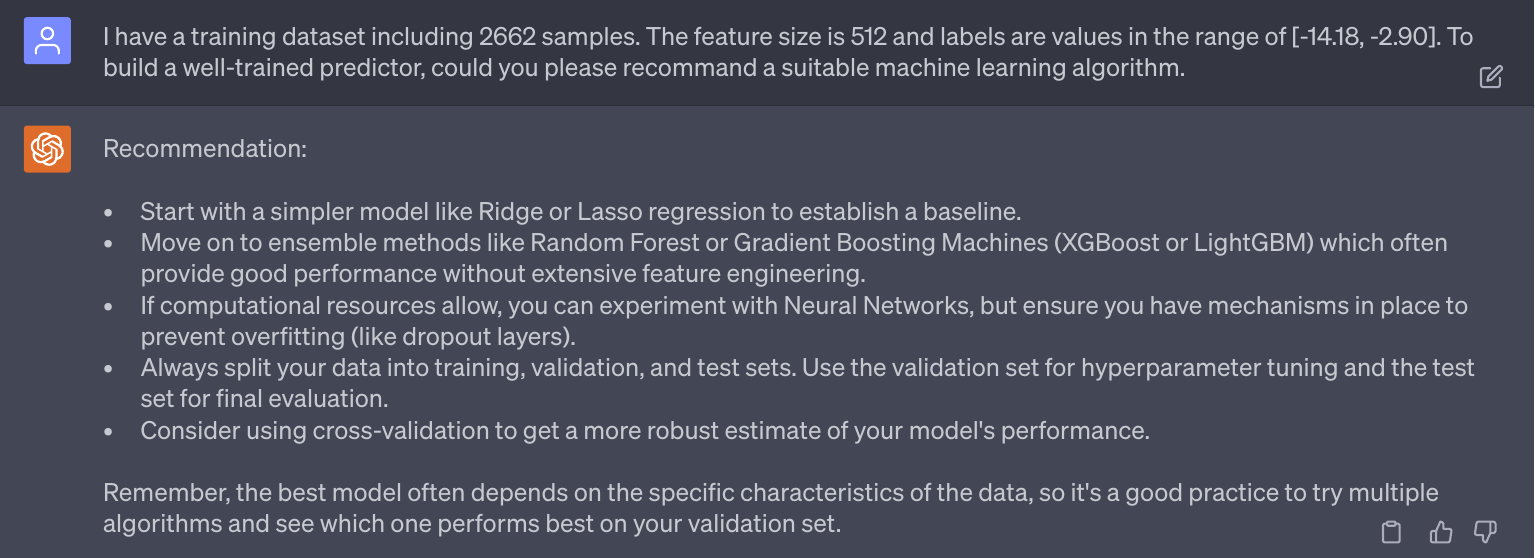}
    \centering
    \caption[name={Graph}]{Consideration of machine learning algorithms that can be applied to build binding affinity predictors.}
    \label{dig:predictors}
\end{dialogue}

\begin{table}[ht]
    \centering
    \setlength\tabcolsep{10pt}
    \caption{Dataset summary. Four datasets are utilized, each containing SMILES strings of inhibitors targeting DAT, NET, SERT, and hERG respectively. Alongside each SMILES string, the respective binding affinity in the unit of kcal/mol is also included as the label for each sample. Additionally, the final two columns represent the 10-fold cross validation Pearson correlation coefficient (R) and root-mean-square rrror (RMSE) for each binding affinity predictor across the four datasets. }
    \begin{tabular}{ccccc}
        \toprule
        Dataset name & Sample size & Binding affinity range (kcal/mol)  & 10-fold R & 10-fold RMSE\\
        \midrule
        DAT-Inhibitors     & 2662         & [-14.18, -2.90]                        & 0.8212      & 0.8979\\
        NET-Inhibitors      & 2981         & [-14.63, -5.47]                        & 0.7732      & 0.9683\\
        SERT-Inhibitors     & 4341         & [-15.00, -5.64]                        & 0.8022      & 0.9448\\
        hERG-Inhibitors     & 6298         & [-13.84, -3.27]                        & 0.8092      & 0.7981\\
        \bottomrule
    \end{tabular}
    \label{tab:dataset}
\end{table}

\begin{figure}[htbp!]
    \includegraphics[width=1.0\textwidth]{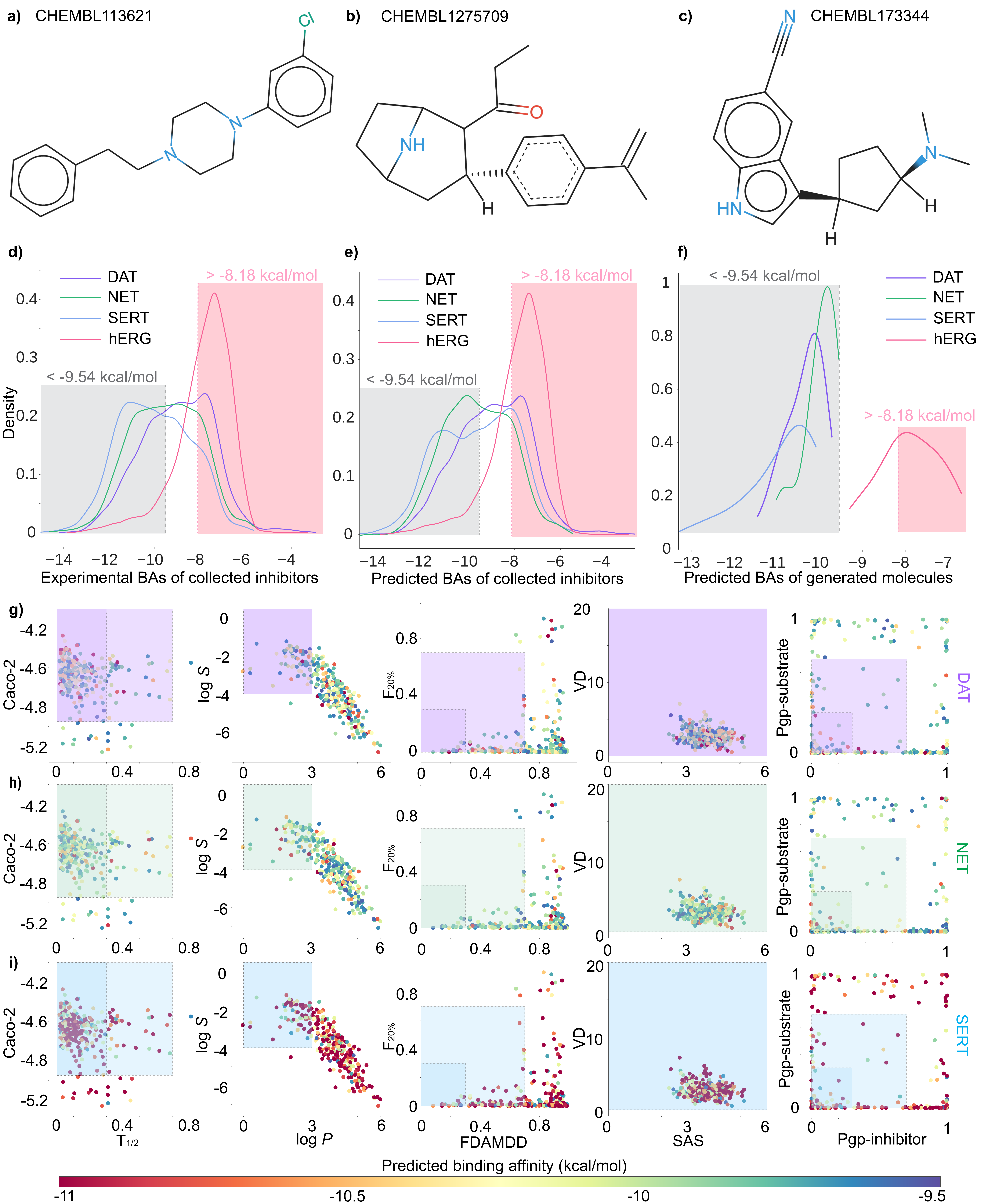}
    \centering
    \caption{2D molecular structures of reference compound with ChEMBL ID {\bf a)} CHEMBL113621 from DAT-Inhibitors dataset, {\bf b)} CHEMBL1275709 from NET-Inhibitors dataset, and {\bf c)} CHEMBL173344 from SERT-Inhibitors dataset. 2D molecular structures are rendered by an online software \href{https://doc.gdb.tools/smilesDrawer/sd/example/index_light.html}{SmilesDrawer 2.0} \cite{probst2018smilesdrawer}. {\bf d)} Distribution of experimental binding affinities for the four training datasets (DAT-Inhibitors, NET-Inhibitors, SERT-Inhibitors , and hERG-Inhibitors). {\bf e)} Distribution of predicted binding affinities derived from the four deep neural network predictors. {\bf f)} Distribution of predicted binding affinities for newly generated inhibitors targeting DAT, NET, SERT, and hERG. {\bf g)} Screening of 330 preliminary multi-target drug candidates. The color of each point represents the predicted binding affinities to DAT (purple,  {\bf g)}), NET (green,  {\bf h)}), and SERT (blue,  {\bf i)}). The light purple, green, and blue frames outline the medium ranges of 10 ADMET, physicochemical, and medicinal chemistry properties, respectively, while the dark purple, dark green, and dark blue frames outline the excellent ranges of these properties.}
    \label{fig:combine 1}
\end{figure}

\autoref{fig:combine 1} {\bf d)} and {\bf e)} shows the experimental and predicted binding affinity distribution on the four training sets: DAT{-Inhibitors}, NET{-Inhibitors}, SERT{-Inhibitors}, and hERG{-Inhibitors}. The distribution of predicted binding affinities align well with the experimental values, which shows that our binding affinity predictor is reliable. The grey region {represents} the zone where the binding affinity is less than -9.54 kcal/mol (i.e $K_{\rm i}=0.1 \mu$M). This value is generally considered as the cut-off for recognizing active compounds. The pink region highlights the zone where the binding affinity is more than -8.18 kcal/mol (i.e $K_{\rm i}=1 \mu$M), a criterion set to prevent hERG-related side effects. We generated around 16 million novel compounds from stochastic-based molecular generator, and their distribution of predicted binding affinities are depicted in \autoref{fig:combine 1} {\bf f)}. It is worth mentioning that the predicted binding affinity of newly generated molecules, targeted to DAT, NET, and SERT, all fall below -9.54 kcal/mol. This results from the properly adjusted hyperparameters and threshold of step size in the stochastic-based molecular generator, aimed at producing more active compounds. However, as hERG inhibitors were not included as reference compounds in the generation of new molecules, only about half of the generated compounds meet the criteria set to prevent hERG-related side effects.


\subsubsection{ChatGPT assisted virtual screening of multi-target drug candidates}\label{subsubsec:screening}
By editing the latent space vector of the Seq2Seq AutoEncoder (AE), we were able to generate a vast number of vectors (around 16 billion) using our stochastic-based molecular generator. These vectors are then decoded into molecules through the GRU Decoder of the Seq2Seq AE. Next, we proceeded by implementing a filtering process {in } which we removed any duplicated molecules and predicted {the} corresponding binding affinities { of remaining molecules} to four target proteins: DAT, NET, SERT, and hERG. Any generated molecules meeting the binding affinity requirement (i.e., $\Delta G < $  -9.54 kcal/mol for DAT, NET, SERT and $\Delta G > $  -8.18 kcal/mol for hERG) were considered preliminary multi-target drug candidates. A total of 330 preliminary drug candidates pass the filtering test. Moreover, the similarities between 330 preliminary drug candidates and three references compounds are all less than 0.5, indicating the the high novelties of generated multi-target molecules. 

{Then}, we sought advice from GPT-4's 1st persona on criteria for selecting drug-like lead compounds. Due to paper length constraints, a concise version of responses can be found in \autoref{dig:druglikeness}. Acting on these suggestions, we utilized in silico tools to predict the Absorption, Distribution, Metabolism, Excretion, and Toxicity (ADMET) properties of each candidate molecule. Specifically, we examined 10 properties of 330 preliminary multi-target drug candidates through \href{https://admetmesh.scbdd.com/docs/#/screening}{ADMETlab 2.0}. This platform aims to provide systematic evaluation of ADMET properties, physicochemical properties, and an assessment of medicinal chemistry friendliness. 

The 10 properties assessed in this work included:  Caco-2 (the human colon adenocarcinoma cell lines) permeability, $\text{F}_{20\%}$ (the human oral bioavailability 20\%), Pgp-substrate (the substrate of P-glycoprotein), Pgp-inhibitor (the inhibitor of P-glycoprotein), VD (volumn density), $\text{T}_{1/2}$ (The half-life of a drug), FDAMDD (The maximum recommended daily dose), SAS (synthetic accessibility score), log $P$ (the logarithm of the n-octanol/water distribution coefficient), and log $S$ (the logarithm of aqueous solubility value). The optimal range of 10 properties can be found in \autoref{tab:optimal range}. 

\begin{dialogue}[ht!]
    \includegraphics[width=1.0\textwidth]{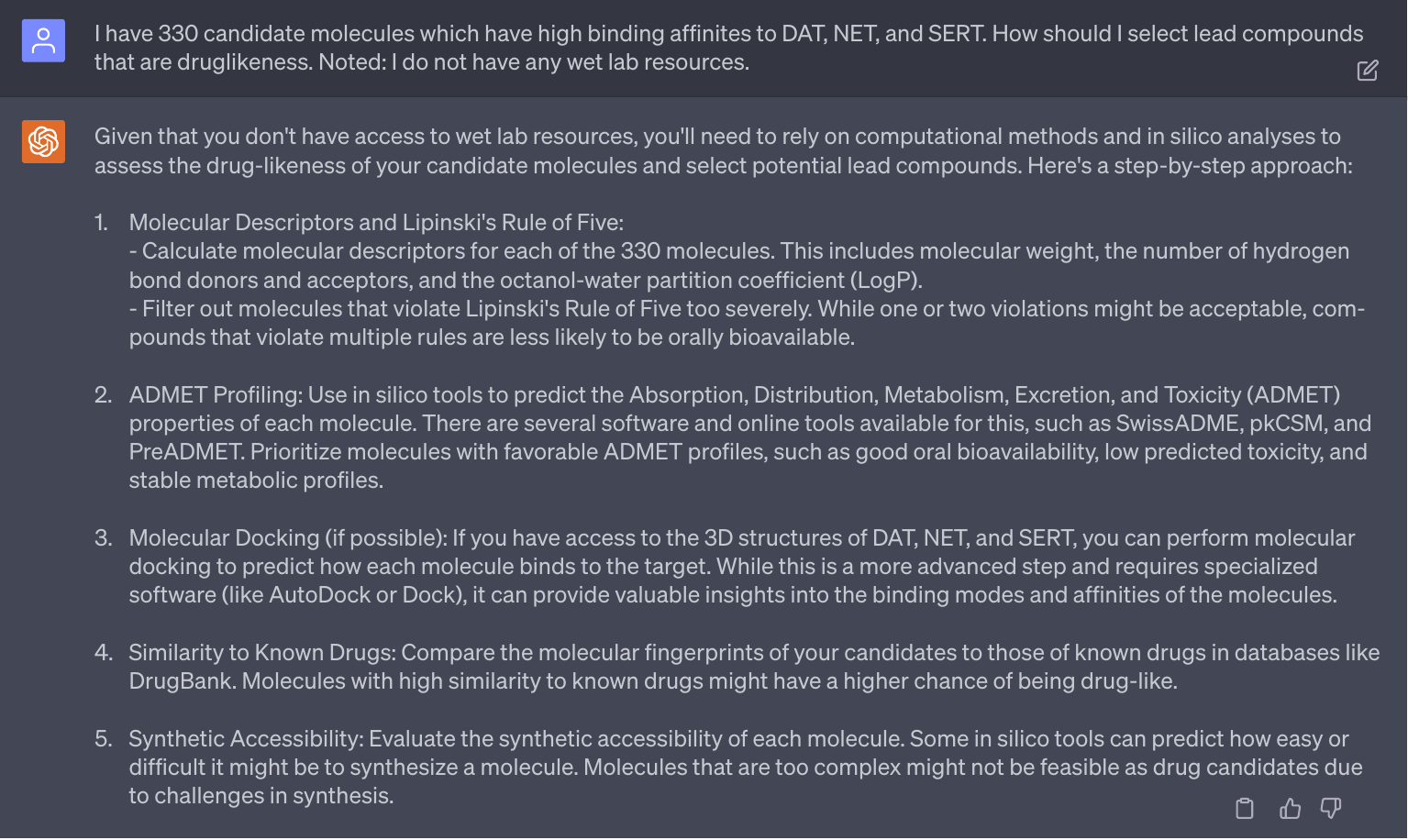}
    \centering
    \caption[name={Graph}]{Silico analyses to assess the drug-likeness of candidate molecules suggested by ChatGPT.}
    \label{dig:druglikeness}
\end{dialogue}


\autoref{fig:combine 1} {\bf g)}, {\bf h)}, and {\bf i)} depict the screening results on 330 preliminary multi-target drug candidates. The color gradient in each panel signifies the predicted binding affinities of the molecules to their respective targets. Specifically, in the \autoref{fig:combine 1} {\bf g)}, the color of each point indicates the binding affinities to DAT. Similarly, in the \autoref{fig:combine 1} {\bf h)} and {\bf i)}, the colors of points represent the binding affinities to NET and SERT, respectively. It can be seen that the binding affinity of drug candidates for SERT is stronger than that for DAT and NET. The frames outline the medium (light purple, green, and blue) and excellent ranges (dark purple, green, and blue) for the 10 evaluated ADMET, physicochemical, and medicinal chemistry properties. Researchers can access the code of scatter plot in python via ChatGPT swiftly { see Supporting Information S4.3}.

\autoref{fig:combine 1} {\bf g)}, {\bf h)}, and {\bf i)} indicate that all the drug candidates have favorable volume density (VD) and synthetic accessibility score (SAS) values. However, only a select few { drug candidates} demonstrate preferable FDAMMDD, $\text{F}_{20\%}$, $\text{T}_{1/2}$, Pgp-sub, and Pgp-inh. Among all, 15 candidate drugs fall within the medium range for all properties, thus are {considered as} potential multi-target anti-cocaine lead compounds. We { also} evaluated the SMILES strings of 15 potential anti-cocaine addiction lead compounds that could target multiple transporters: DAT, NET, and SERT. Notably, all 15 lead compounds satify Lipinski's rule of five.  {  \autoref{subsubsection:molecular interactions}. Furthermore, we also did molecular docking analysis of 15 lead compounds following the suggestions of ChatGPT, which can be found in  \autoref{subsubsection:molecular interactions}.}

Noted, in \autoref{subsubsec: Idea generation}, GPT-4 introduced a generalized suggestion about leveraging alternative machine learning approaches to predict effectiveness of a given molecule against various targets. A detailed discussion of potential ML methods is provided in \autoref{dig:alternative ML}. In fact, ADMETlab2.0 serves as an exemplary application of these alternative methodologies. Specifically, it incorporates both Graph Neural Networks (GNNs) and multitask learning, leading to enhanced performance across many modeled endpoints.

\begin{dialogue}[ht!]
    \includegraphics[width=1.0\textwidth]{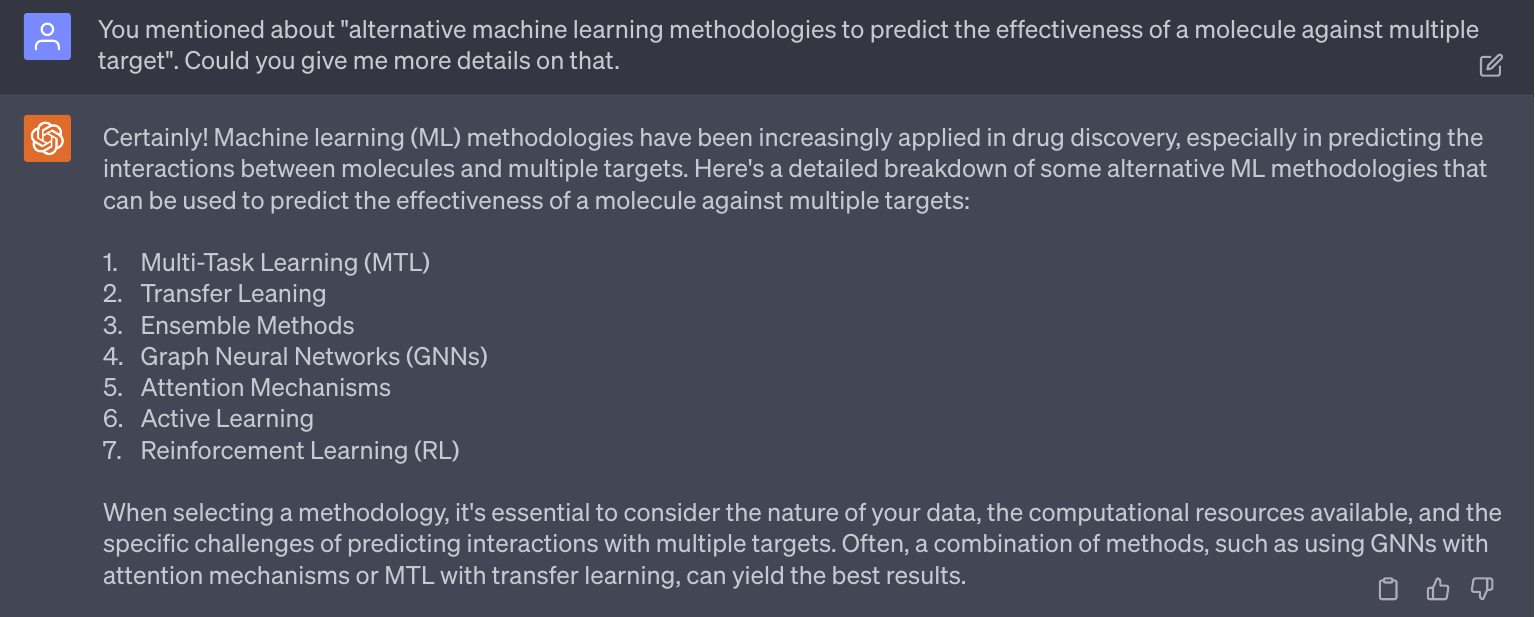}
    \centering
    \caption[name={Graph}]{Alternative machine learning approaches to predict effectiveness of a given molecule against various targets.}
    \label{dig:alternative ML}
\end{dialogue}

\subsubsection{ChatGPT assisted identification of potential multi-target leads}

\begin{figure}[ht!]
    \includegraphics[width=1.0\textwidth]{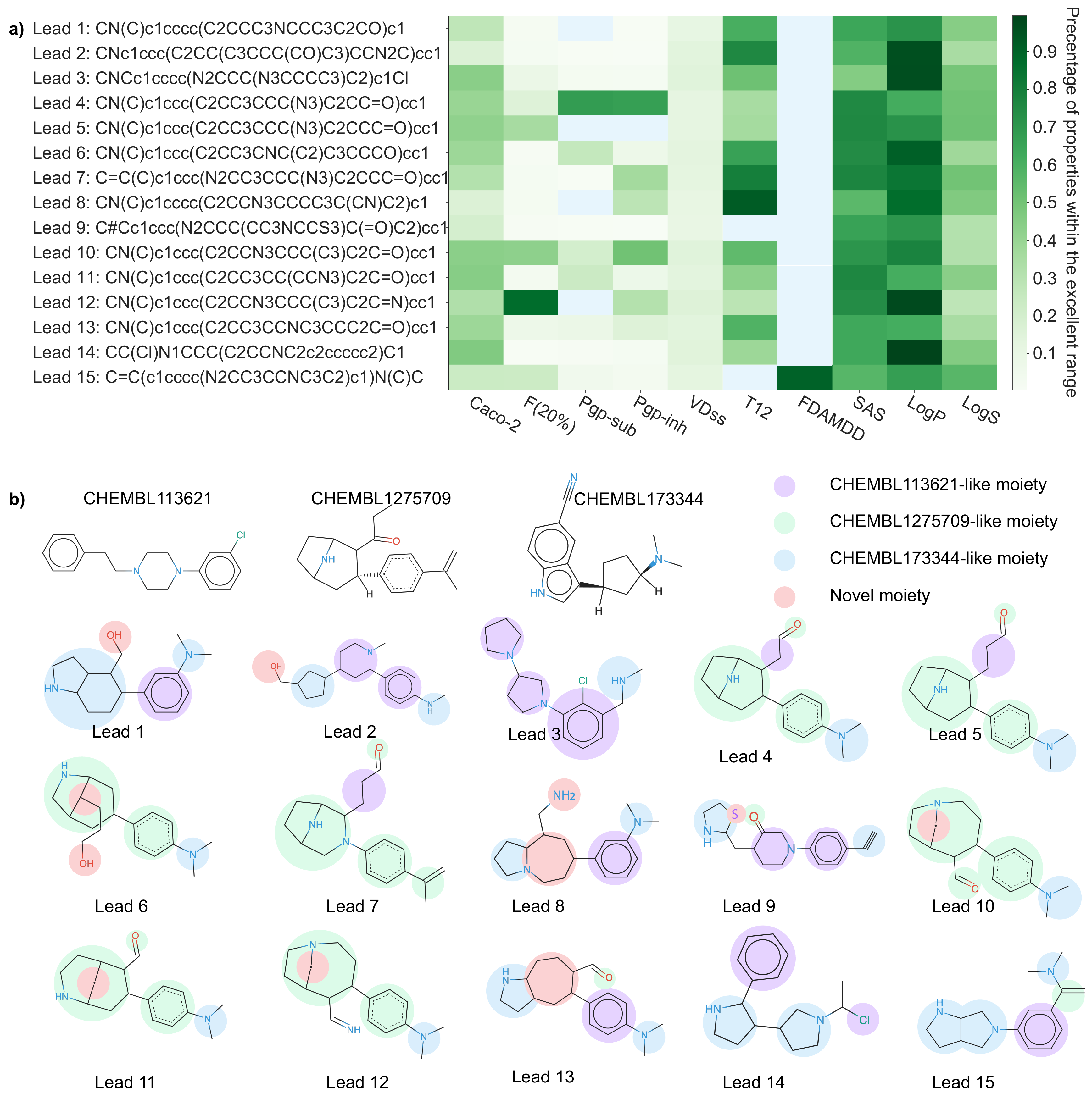}
    \centering
    \caption{{\bf a)} The SMILE string of 15 potential anti-cocaine lead compounds that could target to multiple transporters DAT, NET, and SERT. Green pixels indicate a given candidate falls within the excellent range for each of the 10 evaluated ADMET, physicochemical, and medicinal chemistry properties, while the blue pixels describe a give candidate drug only falls within the medium range of these properties. The color gradient represents the percentage of properties within the excellent range for each given compound. {\bf b)} Illustration of the 2D molecular structures of three reference compounds and 15 potential anti-cocaine lead compounds, which may target multiple transporters (DAT, NET, and SERT). Purple, green, and blue spots represent the CHEMBL113621-like, CHEMBL1275709-like, and CHEMBL173344-like moieties, respectively. Red spots highlight novel moieties that are not present in the three reference compounds. All 2D molecular structures are rendered by an online software \href{https://doc.gdb.tools/smilesDrawer/sd/example/index_light.html}{SmilesDrawer 2.0} \cite{probst2018smilesdrawer}.}
    \label{fig:combine 2}
\end{figure}

\autoref{fig:combine 2} {\bf a)} represents the SMILES strings of the 15 candidate drugs, detailing whether each drug falls within the excellent or medium range for the evaluated 10 properties. Green pixels denote a candidate drug that meets the excellent range for each of the 10 evaluated properties, while {blue} pixels indicate a drug that only achieves the medium range. The color gradient signifies the percentage of properties within the excellent range for each compound. The $y$-axis displays the SMILES strings of the 15 candidate drugs along with their corresponding IDs. Notably, Drug 15 is the {only} candidate that exhibits an excellent FDAMDD value, though its $\text{T}_{1/2}$ is relegated to the medium range. Furthermore, the drugs with IDs 4, 7, 10, and 15 demonstrate a relatively high percentage of properties within the excellent range.


In addition, we aimed to analyze the moieties of 15 potential multi-target anti-cocaine addiction lead compounds. \autoref{fig:combine 2} {\bf b)} depicts the 2D molecular structures of three reference compounds (CHEMBL113621, CHEMBL1275709, and CHEMBL173344) and 15 potential anti-cocaine lead compounds. In this figure, purple, green, and blue spots depict the CHEMBL113621-like, CHEMBL1275709-like, and CHEMBL173344-like moieties, respectively. Additionally, red spots highlight novel moieties that are not present in the three reference compounds. Leads 1, 2, 3, 8, 9, 13, 14, and 15 share more CHEMBL113621-like and CHEMBL173344-like moieties, while Leads 4, 5, 6, 7, 10, 11, and 12 sahre more CHEMBL1275709-like moieties. For a systematic examination of the functional groups within these lead compounds, we leaned on guidance from ChatGPT, which simulated the role of a chemist skilled in interpreting SMILES (Simplified Molecular Input Line Entry System) strings. An illustrative example of this guidance can be seen in \autoref{dig:chemist}. A comprehensive analysis is available in the Supporting Information S4.4. Assisted by ChatGPT, we successfully undertook the analysis of the 2D molecular structures of the 15 lead compounds. 

\begin{dialogue}[ht!]
    \includegraphics[width=1.0\textwidth]{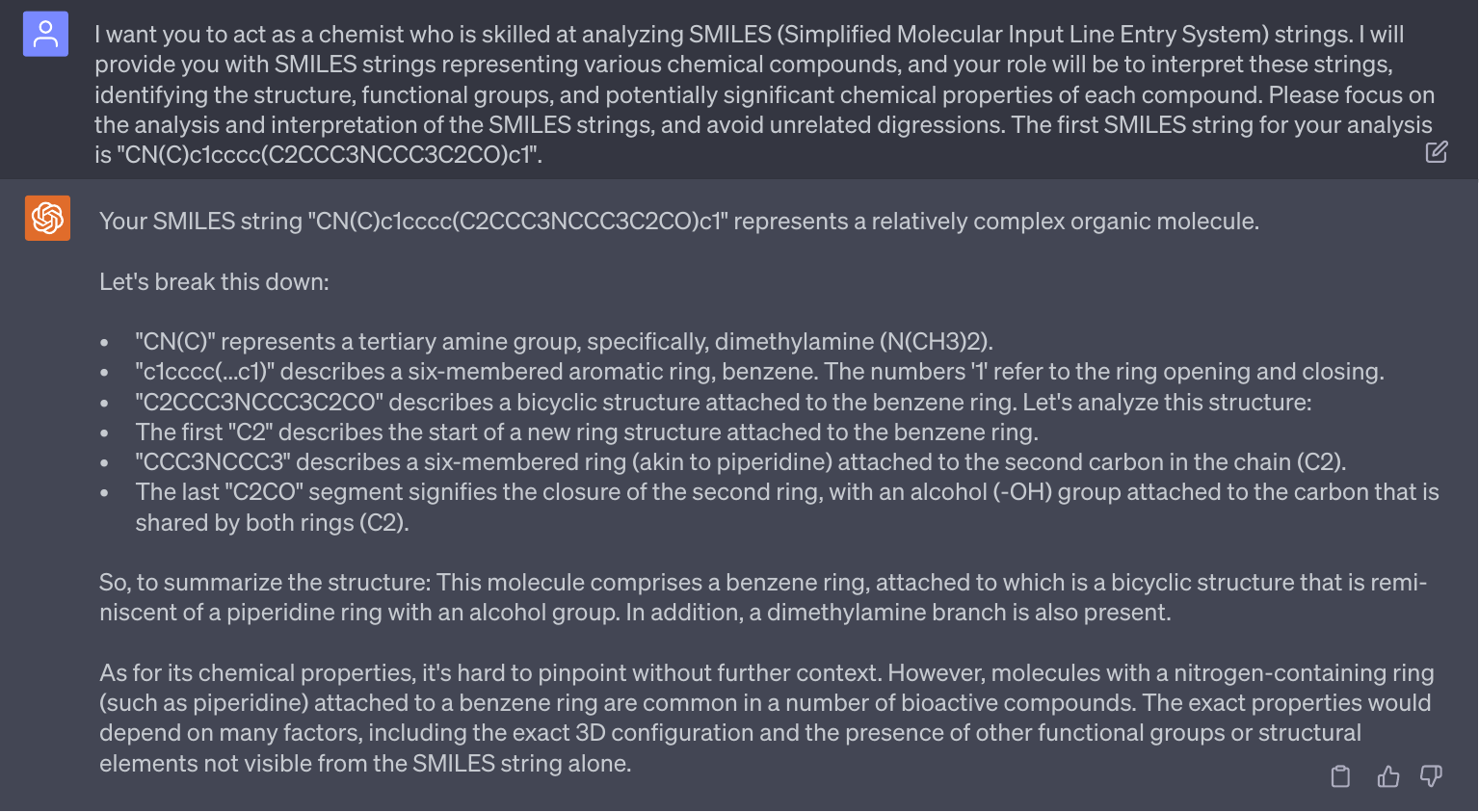}
    \centering
    \caption[name={Graph}]{ChatGPT acts as a chemist to guide the analyze of functional groups. Cpmplete interactions with ChatGPT can be found in Supporting Information S4.4.}
    \label{dig:chemist}
\end{dialogue}

Lead 1 features a benzene ring linked to a bicyclic structure and a dimethylamine branch. It is worth mentioning that molecules with a nitrogen-containing ring (such as piperidine) attached to a benzene ring are frequently found in bioactive compounds.  Lead 2 comprises a benzene ring attached to a modified piperazine ring, which is further connected with a cyclopentane group.. This type of structure is prevalent in numerous bioactive molecules, including some pharmaceutical drugs.  Lead 3 contains a chlorobenzene ring coupled with a substituted piperazine ring. Besides, there is a methylamine group attached to the benzene ring. This structure might have potential psychoactivity, as structures featuring a nitrogen-containing ring connected to a benzene ring are commonly observed in many psychoactive compounds such as Phenethylamines, Tryptamines, and Ergolines.  Lead 4 encompasses a benzene ring with an attached dimethylamine group. In addition, the benzene ring is linked to a bicyclic structure that includes a piperidine ring and an aldehyde group. This molecule could potentially be bioactive due to the presence of both a benzene ring and a nitrogen-containing ring.  Lead 5 shares very similar structure as Lead 4. The only difference is that the bicyclic structure of Lead 5 includes propionaldehyde group instead of aldehyde group. 

Leads 6, 10, 11, and 12 all feature a benzene ring connected to a dimethylamine group and a bicyclic structure. Lead 7 consists of a benzene ring linked to a substituted alkene group, along with a bicyclic structure that includes both a pyrrolidine ring and a piperazine ring. Additionally, this bicyclic structure is connected by a propionadehyde group. Lead 8 includes a benzene ring with an attached dimethylamine group, connected to a bicyclic structure that incorporates a piperidine ring and an additional pyrrolidine ring. Lead 9 comprises a benzene ring linked to an alkyne group, and a complex structure with a piperidine ring and a three-membered nitrogen-sulfur ring. Notably, molecules with sulfur-containing rings, such as penicillin and angiotensin-converting enzyme (ACE) inhibitors, are recognized as bioactive. Lead 13 incorporates a benzene ring with an attached dimethylamine group, connected to a bicyclic structure that includes pyrrolidine ring and a cyclohexane ring which is also connected by a formyl group. Lead 14 is composed of a chloroethane group linked to two pyrrolidine rings and a benzene ring. Finally, Lead 15 consists of a benzene ring linked to a dimethylamine group via an alkene group and attached to a bicyclic structures composed by two pyrrolidine rings.


\subsubsection{ChatGPT assisted analysis  of   cocaine transporter  and   inhibitor interactions}\label{subsubsection:molecular interactions}

As mentioned in \autoref{dig:druglikeness}, ChatGPT suggested to perform molecular
docking to predict how each molecule binds to the target. We decided to accept this suggestion as
the understanding of the molecular mechanism of drug-target interactions is vital in identifying effective drug candidates. We also seek the expertise from ChatGPT for the installation guidance of AutoDock Vina \cite{trott2010autodock} and guidance to execute molecular docking procedures (see \autoref{dig:docking}) between 15 lead compounds and target proteins DAT (PDB ID: 4XPA) and SERT (PDB ID: 6DZZ). To be noted, due to the lack of NET structures in the Protein Data Bank, we do not included molecular interaction analysis of candidate leads with NET. Moreover, we want to visualize 2D protein-ligand interaction diagrams, as they offer a streamlined representation of protein-ligand interactions, highlighting crucial residues, hydrogen bonds, and more. ChatGPT recommended several popular software tools for this purpose, including \href{https://www.ebi.ac.uk/thornton-srv/software/LigPlus/}{LigPlot$^{+}$} and \href{https://www.schrodinger.com/products/maestro}{Maestro}. In this work, we chose LigPlot$^{+}$ for our visualization needs.

\begin{dialogue}[ht!]
    \includegraphics[width=1.0\textwidth]{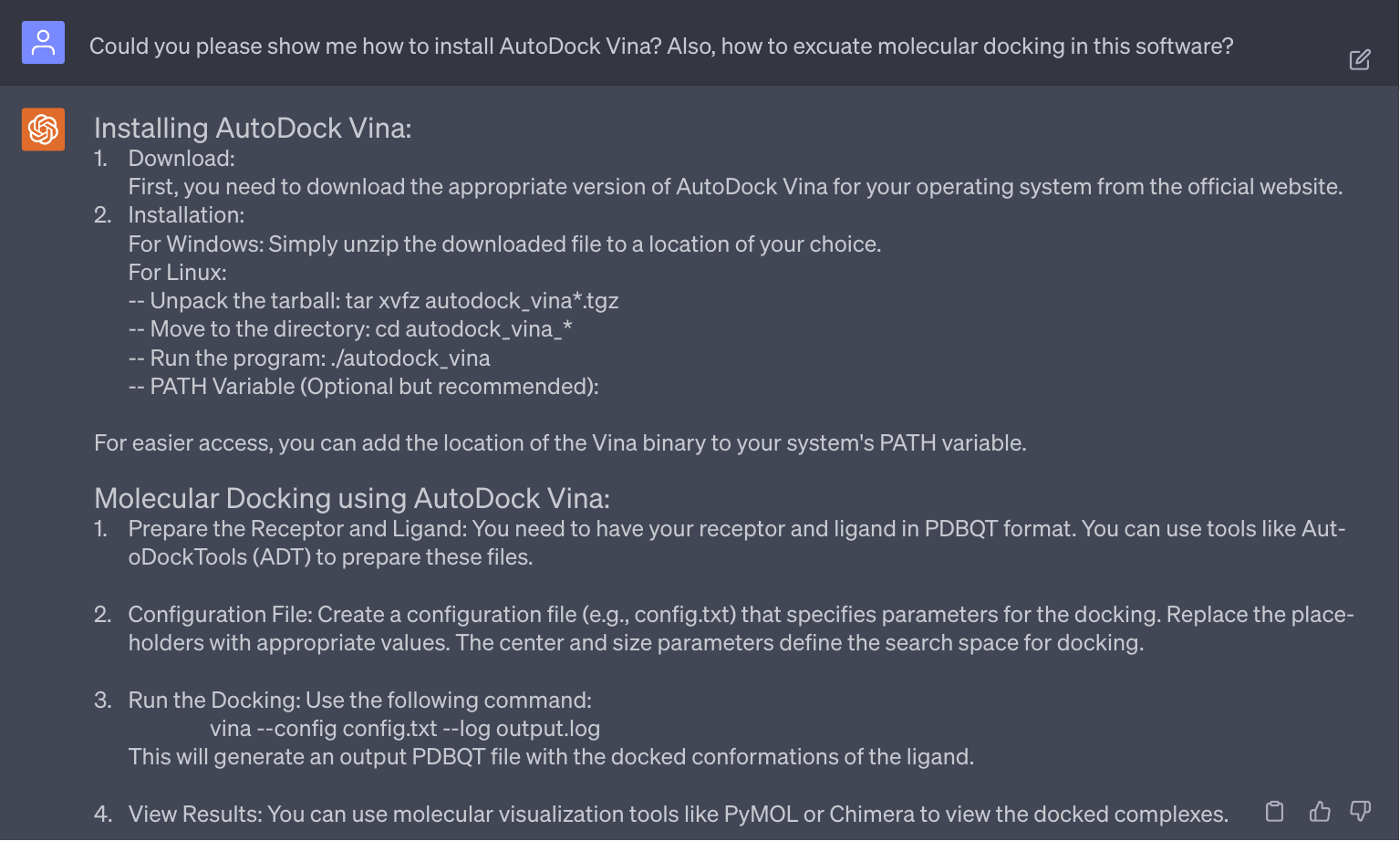}
    \centering
    \caption[name={Graph}]{ChatGPT assists in the software installation and application. Complete dialogue is in the Supporting Information S4.3.}
    \label{dig:docking}
\end{dialogue}

\begin{figure}[ht!]
    \includegraphics[width=1.0\textwidth]{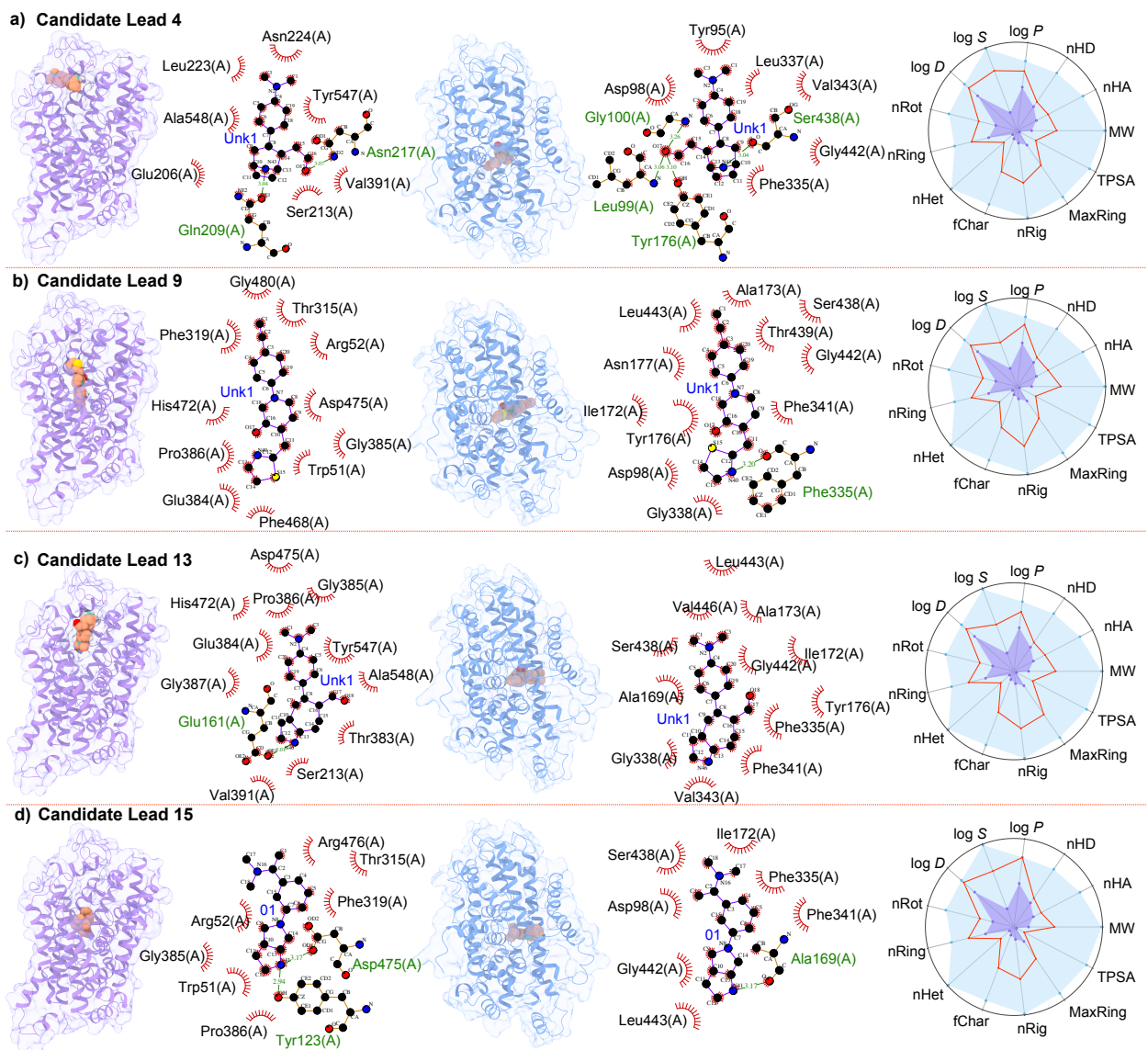}
    \centering
    \caption{Predicted docking poses of selected lead candidates to DAT (1st column) and SERT (3rd column) by AutoDock Vina. DAT is colored in purple, and SERT is presented in blue. The 2nd and 4th columns demonstrate the molecular interaction of these leads with DAT and SERT, respectively. The final column portrays the physicochemical properties of the lead candidates, which include MW (molecular weight), log P (logarithm of octanol/water partition coefficient), log S (logarithm of the aqueous solubility), log D (log P at physiological pH 7.4), nHA (number of hydrogen bond acceptors), nHD (number of hydrogen bond donors), TPSA (topological polar surface area), nRot (number of rotatable bonds), nRing (number of rings), MaxRing (number of atoms in the largest ring), nHet (number of heteroatoms), fChar (formal charge), and nRig (number of rigid bonds). Here the purple dots denote the minimal value and the blue dots indicate the maximal value within the optimal range. The red lines represent the values of the properties for each lead candidate. The figures are categorized as follows: {\bf a)} candidate lead 3, {\bf b)} candidate lead 9, {\bf c)} candidate lead 13, and {\bf d)} Candidate lead 15.}
    \label{fig:combine 3}
\end{figure}

Our observations highlight the critical role of hydrogen bonds in the molecular interactions. The interactions of the drug candidate with DAT and SERT feature two and one hydrogen bonds, respectively, thereby contributing to the high potency of the molecule on the transporters. The first and third columns in \autoref{fig:combine 3} illustrate the docking of Lead 15 and its molecular interactions with DAT and SERT. We have identified 15 nearly optimal leads. As demonstrated in the second and fourth columns of \autoref{fig:combine 3} {\bf a)}, Lead 4 establishes two hydrogen bonds with DAT and four hydrogen bonds with SERT. Of the two bonds with DAT, one is formed between an oxygen atom on the residue Gln209(A) and a nitrogen atom on the compound, and the other involves an oxygen atom in a hydroxyl group of the compound interacting with a nitrogen atom on residue Asn207(A) of DAT. Among the four hydrogen bonds formed between Lead 4 and SERT, two involve the same oxygen atom interacting with a nitrogen atom on residues Leu99(A) and Tyr176(A) of SERT, while the other two bonds are formed by nitrogen atoms on Lead 4, interacting with oxygen atoms on residues Ser438(A) and another unidentified residue of SERT. \autoref{fig:combine 3} {\bf b)} depicts a hydrogen bond in the molecular interactions between candidate Lead 9 and SERT. This bond is formed by a nitrogen atom on the compound and an oxygen atom in a hydroxyl group on residue Phe335(A) of SERT. However, no hydrogen bond is observed in its interactions with DAT. This suggests that other types of interactions, such as hydrophobic bonds, may play a major role in the high binding affinity between Lead 9 and DAT.


The molecular docking poses of Lead 13 on DAT and SERT are illustrated in the 1st and 3rd columns of \autoref{fig:combine 3} {\bf c)}. In the second column of \autoref{fig:combine 3} {\bf c)}, a single hydrogen bond can be observed between a nitrogen atom of Lead 13 and an oxygen atom on the residue Glu161(A) of DAT. Conversely, no hydrogen bond is detected between Lead 13 and SERT, as demonstrated in the 4th column of \autoref{fig:combine 3} {\bf c)}. The molecular docking poses of Lead 15 on DAT and SERT are portrayed in \autoref{fig:combine 3} {\bf d)}, presenting the compound's docking positions at the centers of both transporters. In its interaction with DAT, Lead 15 forms two hydrogen bonds through a nitrogen atom in a five-membered nitrogen heterocycle. This nitrogen atom interacts with oxygen atoms in two hydroxyl groups, which are attached to the residues Asp475(A) and Tyr123(A) of DAT. Moreover, a hydrogen bond exists between the candidate drug Lead 15 and SERT. This bond is formed by the same nitrogen atom in the five-membered nitrogen heterocycle, which interacts with an oxygen atom in a hydroxyl group attached to the residue Ala169(A) of SERT. The molecular interaction with other 11 candidate leads can be found in the Supporting Information S3.



\section{Discussion}

\subsection{Scrutinizing chatbots}
\begin{figure}[ht!]
    \includegraphics[width=1.0\textwidth]{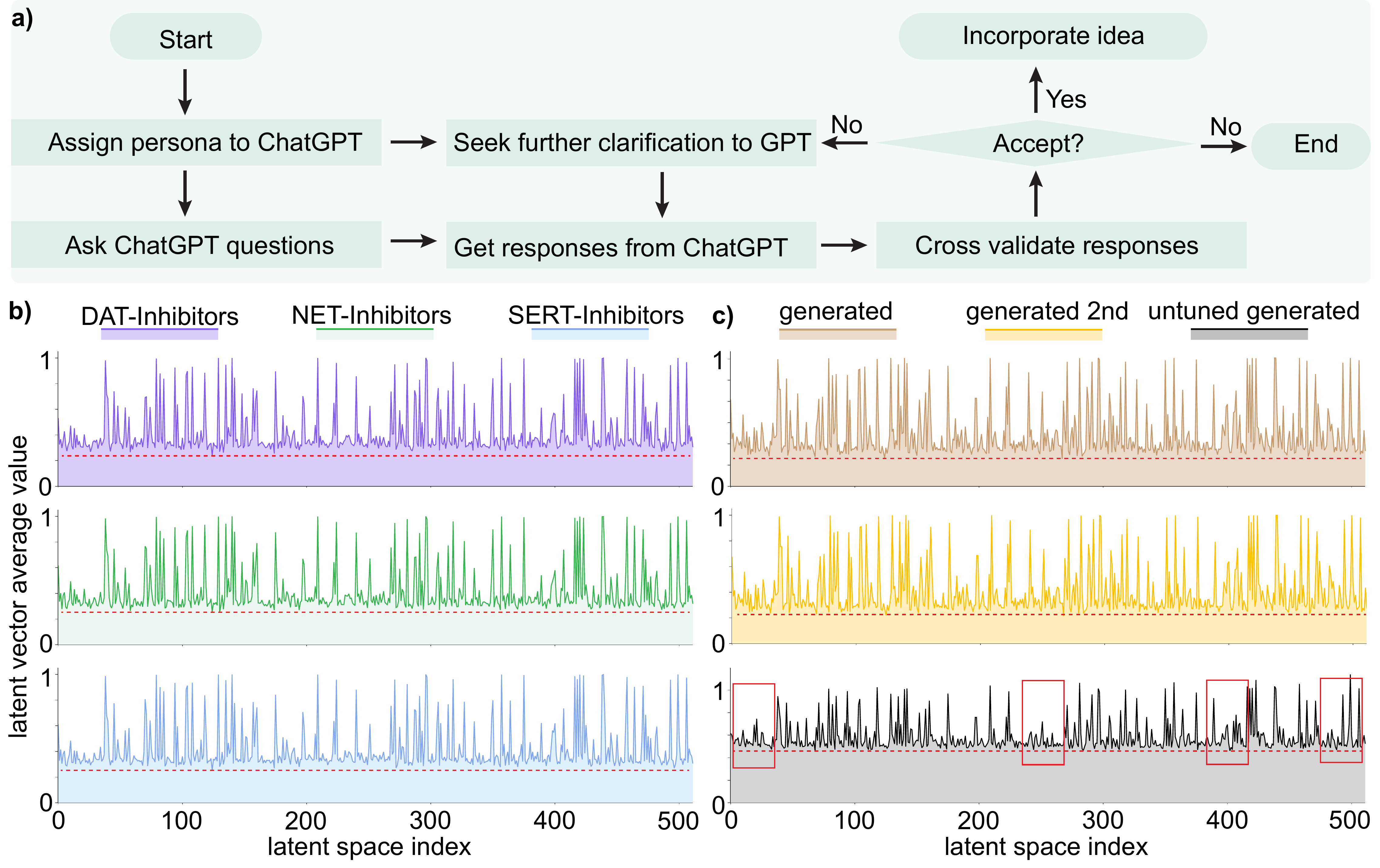}
    \centering
    \caption{{\bf a)} Flowchart of implementing ChatGPT as a virtual guide. {\bf b)} Distribution of latent space vectors across various datasets. The $x$-axis represents the latent space index, which ranges from 0 to 511, while the y-axis denotes the absolute average value of the latent space vector corresponding to each index. The purple, green, and blue figures represent the latent space vector distribution from the DAT-Inhibitors, NET-Inhibitors, and SERT-Inhibitors respectively. {\bf c)}	Distribution of latent space vectors of generated molecules. The brown panel illustrates the latent space vector distribution from generated molecules from the fine-tuned stochastic-based molecular generator. The yellow distribution portrays generated molecules that have been processed by the GNC a second time. The grey distribution corresponds to generated molecules from an untuned stochastic-based molecular generator. }
    \label{fig:combine 4}
\end{figure}

While chatbots are powerful large language models, they are not infallible. Their predictions are heavily reliant on the training data, which may lead to incomplete, outdated, bias, or skewed understandings of certain contexts. Consequently, this could result in the generation of misleading narratives and incorrect information. Therefore, it is essential for researchers to employ chatbots with appropriate care and vigilance. Scientists should not solely rely on chatbots for their research pursuits and should consistently cross-check the information generated by chatbots. Notably, the role of a chatbot like GPT-4 is to assist researchers, not to replace them.  In our current project, we have employed GPT-4 to assist in our anti-cocaine addiction drug discovery process, as delineated in \autoref{fig:combine 4} {\bf a)}.

We first will assign a proper persona to GPT-4 and then ask it with questions. Once we get the response from GPT-4, it is crucial to decide whether to accept the responses or not. If the information aligns well with the literature and our expertise, we will accept the responses and proceed with the suggestions of GPT-4. Otherwise, we will either reject the answer or seek further clarification to GPT to get alternative feedback.  

For example, when acting as a chemist to analyze the functional groups of Leads 2, 7, 8, 9, 13, and 15, ChatGPT provided inaccurate information. Specifically, for Lead 15, ChatGPT identified a structure where a benzene ring is linked to a dimethylamine group via an alkene group and connected to a piperazine ring. However,the dimethylamine group is connected to a piperazine ring instead of a piperazine ring. This misinformation in the interaction with ChatGPT is documented in \autoref{dig:misinfo chemist}. Thus, it is paramount for researchers to verify the accuracy and reliability of responses from ChatGPT using their expertise.

\begin{dialogue}[ht!]
    \includegraphics[width=1.0\textwidth]{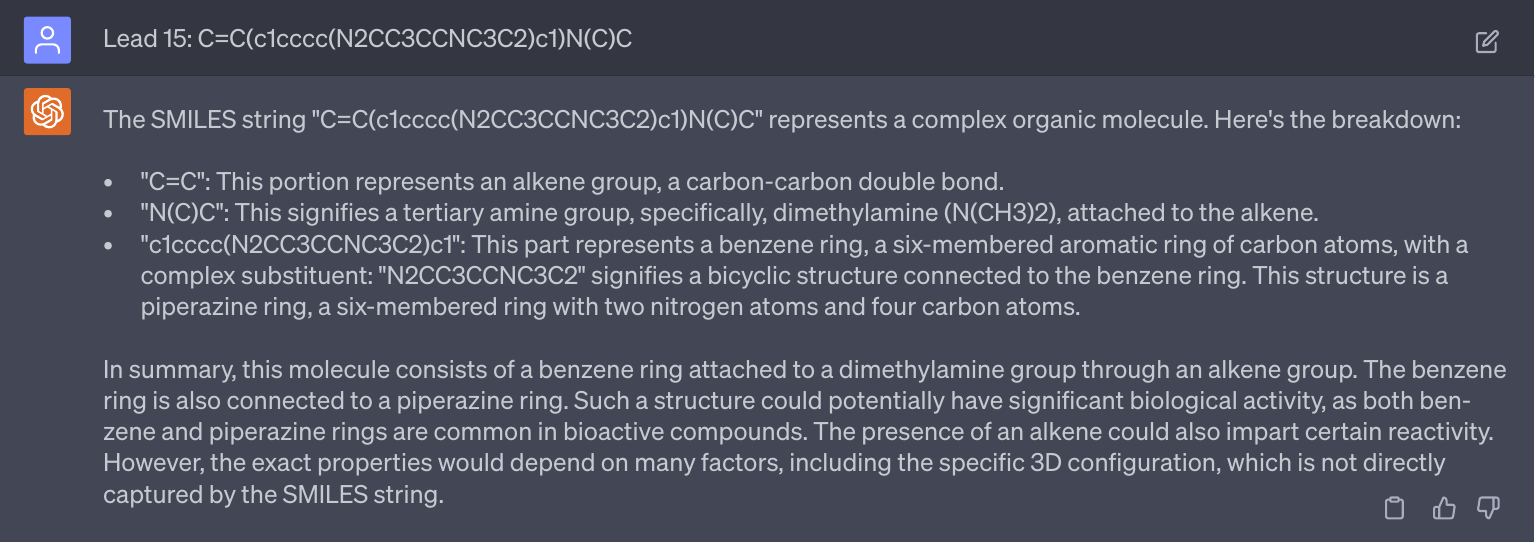}
    \centering
    \caption[name={Graph}]{Misinformation provided by ChatGPT. Complete interactions with ChatGPT can be found in Supporting Information S4.4.}
    \label{dig:misinfo chemist}
\end{dialogue}


Additionally, we noticed that ChatGPT does not perform well when providing methodological explanations. Its responses from the 2nd persona contain some incorrect definitions and explanations. In such cases, we opted not to accept responses from ChatGPT, and seek further clarification followed by the workflow in \autoref{fig:combine 4} {\bf a)}. An effective approach involved pointing out the inaccuracies to ChatGPT and supplying it with accurate references or information, prompting it to adjust its responses. Specific instances of these methodological inaccuracies are detailed in \autoref{dig:misinfo methods}. Despite the inaccurate responses provided by the 2nd persona of ChatGPT, it remained invaluable in helping us grasp numerous theoretical concepts and their interrelations, serving as an effective browser tool.

\begin{dialogue}[ht!]
    \includegraphics[width=1.0\textwidth]{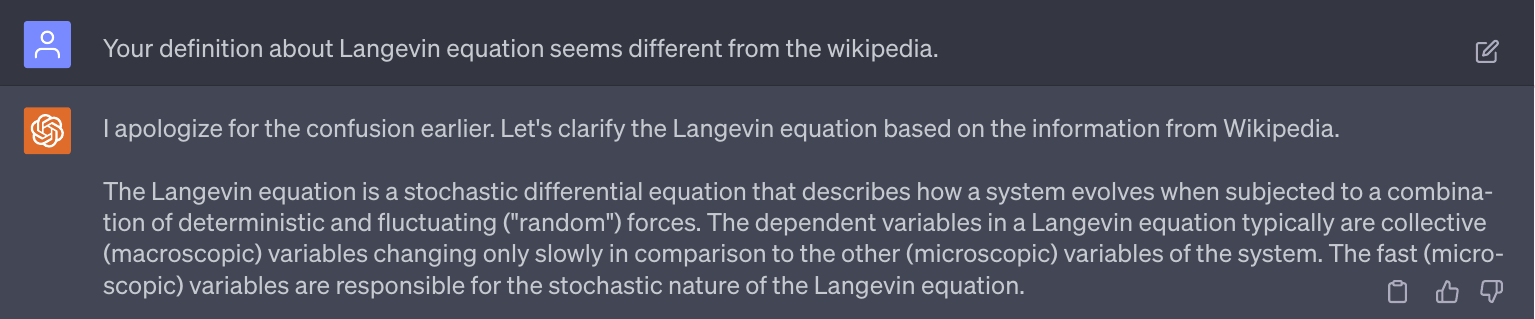}
    \centering
    \caption[name={Graph}]{A succinct dialogue highlighting inaccuracies of ChatGPT responses and suggestions for corrections. Complete interactions with ChatGPT can be found in Supporting Information S4.2.}
    \label{dig:misinfo methods}
\end{dialogue}

\subsection{Autoencoder reconstruction rate of molecules }
A Sequence-to-Sequence Autoencoder (Seq2Seq AE) is a specific type of neural network model designed to learn a compressed representation of input data and reconstruct this data from the obtained representation. The core objective of such an autoencoder is to minimize the discrepancy between its input and output data. In this study, we initially fed the Seq2Seq AE with SMILES strings derived from four distinct datasets to examine their respective reconstruction rates. The calculated reconstruction rates for DAT-Inhibitors, NET-Inhibitors, SERT-Inhibitors, and hERG-Inhibitors datasets are 0.958, 0.970, 0.968, and 0.950 respectively. These values signify a successfully implemented autoencoder model.

In addition to the aforementioned, we verified the reconstruction rate of molecules generated via the Seq2Seq AE. After eliminating duplicated SMILES strings from the generated set, the resulting reconstruction rate stood at 0.996. This high reconstruction rate implies that the distribution of our generated molecules closely mirrors that of the original dataset processed by Seq2Seq AE, underscoring the reliability of the molecules generated by our method.

\subsection{Patterns sensitive latent space vector distributions}
Initially, we introduced random Gaussian noise, with a range from -1 to 1, into the stochastic-based molecular generator. However, these perturbations in the latent space vectors resulted in  weird SMILE strings once decoded. Seeking guidance, we consulted ChatGPT as referenced in \autoref{dig:distribution}, which provided us with eight potential solutions. After reviewing these suggestions, we aligned with the first and second suggestions that echoed findings from our previous work, emphasizing that random perturbations in the latent space can destabilize the Seq2Seq AE model \cite{gao2020generative}. To ensure the reliability and effectiveness of the decoder in Seq2Seq AE model, it is essential to maintain a similar distribution pattern between the original latent space vectors and those derived from the stochastic-based molecular generator.
Therefore, we take efforts to tune the noise that added to the stochatic-based molecular generator, to guarantee the modified/edited latent space vectors retain a representation that the GNC model has learned to decode effectively. 

\autoref{fig:combine 4} {\bf b)} and {\bf c)} depict distribution of latent space vectors across various datasets. Here, the $x$-axis represents the latent space index (ranges from 0 to 511), and the $y$-axis shows the absolute average value of the latent space vector corresponding to each index. The representation of absolute average value helps in visualizing the pattern and magnitude of latent space vectors across a broad index range. The purple, green, and blue panels depict the latent space vector distribution from the DAT-Inhibitors, NET-Inhibitors, and SERT-Inhibitors respectively. The discrepancy can be observed in the grey panel of \autoref{fig:combine 4} {\bf c)} (particularly in the red boxed area). None of the molecules generated by this untuned stochastic-based molecular generator passed either the binding affinity requirements or ADMET tests. Subsequently, we adjusted the Gaussian noise in the stochastic-based molecular generator to ensure the edited latent spaces (represented in brown) exhibited a similar distribution to the original ones as shown in \autoref{fig:combine 4} {\bf b)}. This controlled noise (ranges from -0.1 to 0.1) proved beneficial, leading to the generation of 15 promising leads capable of targeting DAT, NET, and SERT. Additionally, enlightened by ChatGPT (suggestion 8 in \autoref{dig:distribution}), we also implemented a feedback loop where the molecules generated by our fine-tuned stochastic molecular generator is re-encoded into the latent space. It is worth noting that these re-encoded latent space vectors maintained a similar distribution, suggesting that the modifications made to the latent vectors are within the learned parameters of the SGNC.

\begin{dialogue}[ht!]
    \includegraphics[width=1.0\textwidth]{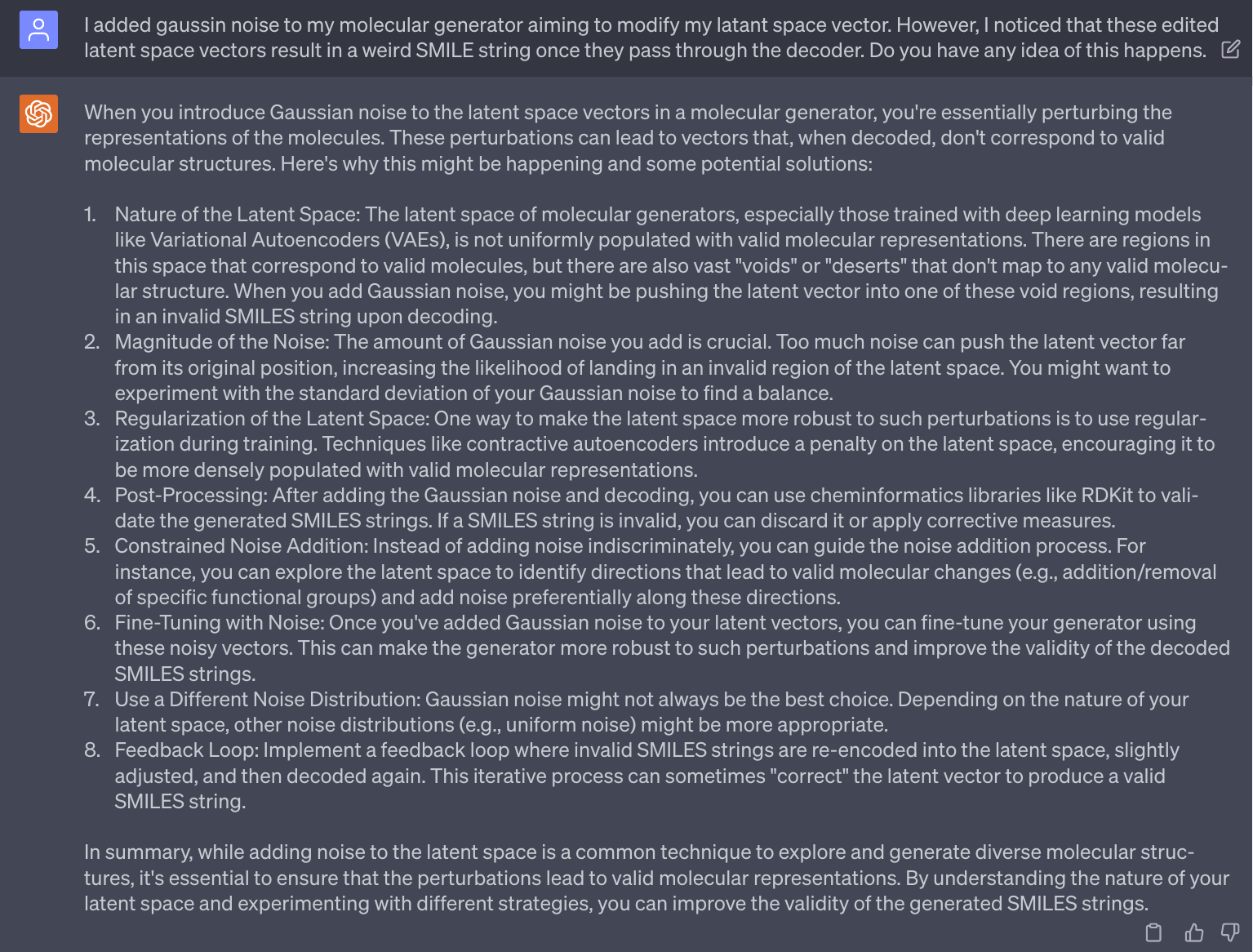}
    \centering
    \caption[name={Graph}]{Solutions provided by ChatGPT regarding the random perturbation issue in the stochastic-based molecular generator.}
    \label{dig:distribution}
\end{dialogue}



\section{Methods }
\subsection{Datasets preparation}\label{subsec:datasets preparation}
Four pharmaceutical targets are key to treating cocaine addiction and drug discovery: Dopamine Transporter (DAT), Norepinephrine Transporter (NET), Serotonin Transporter (SERT), and Human Ether\'a-go-go-Related Gene (hERG). DAT is responsible for dopamine reuptake from synapses back into neurons, terminating neurotransmitter signaling. This causes dopamine accumulation in synapses and inducing intense euphoria. Similarly, NET is inhibited by cocaine, leading to elevated norepinephrine levels in synapses, contributing to  stimulant effects. Furthermore, the inhibition of cocaine and SERT increases serotonin levels in the synapse, resulting in mood elevation, anxiety, and paranoia. Thus, a compound that concurrently modulates DAT, NET, and SERT activities could potentially treat cocaine addiction. Additionally, blocking the hERG potassium ion channel can lead to potentially fatal cardiac arrhythmias. Therefore, it is also critical to consider binding affinity between hERG and new generated leads.

In this study, we collected SMILES strings and binding affinities of inhibitors targeting DAT, NET, SERT, and hERG from the ChEMBL database. A summary of the size and label of datasets can be found in \autoref{tab:dataset}. We leveraged these four datasets in three ways: 1) training four separate binding affinity predictors based on the DAT-Inhibitors, NET-Inhibitors, SERT-Inhibitors, and hERG-Inhibitors datasets respectively, 2) selecting molecules from the DAT-Inhibitors, NET-Inhibitors, and SERT-Inhibitors datasets as reference and seed compounds for the generation of new drug-like molecules, and 3) inputting the SMILES strings from all four datasets into a sequence-to-sequence AutoEncoder model to validate their successful reconstruction rate.

To be noted, we use the binding affinity in terms of Gibbs free energy ($\Delta G$) with units of kcal/mol, rather than the inhibition constant ($K_{\rm i}$) or half maximal inhibitory concentration (IC$_{50}$). The Gibbs free energy of binding provides a direct measure of the favorability of the binding process, which indicates a more favorable binding interaction that can be interpret in the drug-target interactions. The conversion from $K_{\rm i} $ to $\Delta G$ is \cite{cang2018integration}:
\[
\Delta G = - RT \text{ ln} K_{\rm i} \approx 1.3633\times \log_{10} K_{\rm i} \text{ (kcal/mol)},
\]
where $R$ is the gas constant (1.987 cal/mol·K) and $T$ (298.15 K) is the absolute temperature. In addition, the IC$_{50}$ can be approximated to $K_{\rm i}$ in the case of competitive and uncompetitive inhibition according to Kalliokoski \cite{kalliokoski2013comparability} under:
\[
K_{\rm i} = \text{IC}_{50}/2.
\]

\subsection{Stochastic-based generative network complex (SGNC)}
In this section, after thorough evaluation and incorporation of  suggestions from GPT-4, we introduce the stochastic-based generative network complex (SGNC) as a novel mathematical-AI model, {which} is designed to generate novel molecules that potentially serve as effective treatments for cocaine addiction. Specifically, these molecules are intended to target multiple sites such as the Dopamine Transporter (DAT), Norepinephrine Transporter (NET), and Serotonin Transporter (SERT).

\autoref{fig:workflow bg} illustrates the workflow of SGNC, which essentially consists of four main components: 1) Sequence-to-Sequence AutoEncoder (shown in green), 2) Binding Affinity Predictors (shown in yellow), 3) Stochastic-based Molecular Generator (shown in blue), and 4) Analysis via ADMET Lab (shown in purple). The dark arrows represent the training process, the brown arrows show the validation process, and the red arrows indicate the generation process. 

For the training process, we leveraged a well-established translation model, specifically a sequence-to-sequence (Seq2Seq) AutoEncoder (AE). This model was developed to map the International Union of Pure and Applied Chemistry (IUPAC) representation of a molecule to its Simplified Molecular Input Line Entry System (SMILES) representation, as mentioned in \cite{winter2019learning}. In our prior research, we modified this model by switching the input from the IUPAC representation of molecules to their corresponding SMILES strings. 

The generation process involves the following main steps:
\begin{enumerate}
    \item We initially selected one molecule each from the DAT{-Inhibitors}, NET{-Inhibitors}, and SERT{-Inhibitors} datasets. These molecules were chosen because of their relatively high similarity within the three datasets, thereby acting as our reference compounds. In addition, we selected a compound known for its potency against all three targets to serve as our seed compound. 
    \item We then put the reference and seed compounds into the pretrained encoder and extracted the corresponding latent vectors from the latent space of the Seq2Seq AE. Subsequently, we modified the seed vector in the stochastic-based molecular generator, using the information from the reference molecules as a guide. As a result, the generator was capable of producing a large number of new latent vectors. These vectors were then decoded into SMILES strings, which are potentially effective against multiple targets, namely DAT, NET, and SERT. 
    \item Furthermore, we put these decoded SMILES strings into our binding affinity (BA) predictors to filter out molecules that meet our BA requirements (i.e., $\Delta G < -9.54$ kcal/mol on DAT, NET, SERT and  $\Delta G > -8.18$ kcal/mol on hERG). 
    \item Finally, we used ADMETlab 2.0 to select drugable molecules from the generated SMILES with desirable BA properties. This final step in the generation process ensures that the compounds not only bind effectively to the desired targets but also have the necessary absorption, distribution, metabolism, excretion, and toxicity (ADMET) properties for a potential lead compound.
\end{enumerate}

During the validation phase, we first input the SMILES strings from the DAT{-Inhibitors}, NET{-Inhibitors}, SERT{-Inhibitors}, and hERG{-Inhibitors} datasets into our well-trained Seq2Seq AE model to obtain decoded SMILES. Successful reconstruction of the input SMILES {indicates} the reliability of the Seq2Seq AE model. Furthermore, we put the generated SMILES, which have been processed through the stochastic-based molecular generator, into the pre-trained Seq2Seq AE model. In case of unsuccessful reconstruction, we adjust the hyperparameters of the stochastic-based molecular generator until a high reconstruction rate is achieved. This process indicates that the latent vectors edited by the stochastic-based molecular generator maintain a similar distribution to the original latent space vectors from the encoder, which further ensures that our SGNC model is capable of generating chemically feasible compounds, reflecting its potential in drug discovery applications.


\subsubsection{Sequence-to-sequence autoencoder}
The Sequence-to-sequence autoencoder (Seq2Seq AE) is an artificial neural network model used for translating the IUPAC representation of a given molecule into its SMILES string representation \cite{winter2019learning}. In our study, the Seq2Seq AE accepts the SMILES representation of a molecule as the input for the encoder. Subsequently, the latent space of the Seq2Seq AE preserves the structural and functional properties of the provided SMILES. This low-dimensional latent space representation can then be processed by the decoder of the Seq2Seq AE to reconstruct the original SMILES representation. Here, the network that used in encoder and decoder is the gated recurrent unit (GRU). In this work, the pretrained Seq2Seq AE model was utilized from a previous work by Winter et al \cite{winter2019learning}.

The Seq2Seq AE model employed in our study was pretrained on 72 million compounds \cite{winter2019learning} sourced from the \href{https://zinc15.docking.org/}{ZINC15} and \href{https://pubchem.ncbi.nlm.nih.gov/}{PubChem} databases. All duplicate entries within these databases were eliminated and subjected to RDKit \cite{landrum2013rdkit} filtering using the following criteria: 1) only organic molecules, 2) molecular weight between 12 and 600 daltons, 3) more than 3 heavy atoms, 4) partition coefficient log $P$ between -5 and 5, 5) sterochemistry was removed, 6) salts were stripped.

\subsubsection{Binding affinity predictors}
We constructed four binding affinity predictors based on four training datasets: DAT-Inhibitors, NET-Inhibitors, SERT-Inhibitors, and hERG-Inhibitors. These predictors are designed to estimate the binding affinity of potential molecules to four critical targets: DAT, NET, SERT, and hERG. The construction of the predictors involved the following steps:
\begin{enumerate}
    \item[1)] Feature extraction: molecular features (or fingerprints), were derived from the latent space of a sequence-to-sequence AutoEncoder (Seq2Seq AE).
    \item[2)] Label assignment: The labels used for model training were the binding affinities of the molecules to their respective targets.
    \item[3)] Model training: we trained the predictors using PyTorch. Each network consisted of three hidden layers with 512, 1024, and 512 neurons, respectively. The networks were trained over 1000 epochs, with a learning rate of 0.0001 for the first 500 epochs and 0.00001 for the remaining 500 epochs. We chose the Adam optimizer and batch size 16 for this task.
\end{enumerate}

\subsubsection{Stochastic-based molecule generator}
Generative models have gained prominence as potent tools for the generation of prospective new leads. Building upon our prior work, we introduced the Generative Network Complex (GNC), a model specifically tailored to produce novel, drug-like molecules \cite{gao2020generative}. To augment the efficacy of the GNC model, and with guidance from GPT-4, we decided to integrate principles from diffusion-based models \cite{xu2022geodiff, corso2022diffdock}. 

The Langevin equation is a stochastic differential equation (SDE) that used to decribe diffusion processes. This equation {equips} the random trajectories of particles in their velocity space, accounting for both deterministic and stochastic forces. A pivotal goal of this research is to employ the Langevin equation suggested by ChatGPT as a mechanism to enhance the molecular generator present in the GNC model.

Assume $\mathbf{X}$ is a latent space vector of a molecule with 512 dimensions, and $\mathbf{X}_k$ represents its $k$-th latent space reference vector. Then the Langevin equation of our drug generator system is:
\begin{equation}
    \frac{d\mathbf{X}}{dt} = \alpha \sum_k a_k(\mathbf{X}_k - \mathbf{X}) + \bm{\xi}(t),
\end{equation}
where $a_k$ is a positive weighting parameter corresponds to $\mathbf{X}_k$ statisfying $\displaystyle \sum_k a_k =1$, $\bm{\xi}(t)$ is a Gaussian white noise, and $\alpha$ is a hyperparameter. Then according to the Langevin equation in the Supporting Information S1.1.4, the general solution of this system is given by:
\begin{equation}\label{eq:1d Langevin eq}
    \mathbf{X}(t) = \mathbf{C}^{-\alpha t} + \int_0^t e^{-\alpha(t-u)} (\alpha\sum_k a_k\mathbf{X}_k + \bm{\xi}(u))du,
\end{equation}
where the initial state $\mathbf{X}(0) = \mathbf{C}$.

While the Langevin equation offers a microscopic depiction of the diffusion process, a comprehensive understanding of the temporal evolution of particle distribution requires a more macroscopic viewpoint. To bridge this gap, we also introduce the Fokker-Planck equation. Derived from the Langevin equation (a detailed derivation is available in Supporting Information S1.1.5), this equation provide the connection between the dynamics of individual particles and the overarching behavior of the entire system.

\subsubsection{Drug screening}
In drug discovery, several criteria are leveraged to filter promising drug candidates. In our work, we consider molecules that fulfill the following requirements as viable drug prospects: 1) exhibit favorable ADMET properties, 2) comply with Lipinski's rule of five, 3) are synthetically accessible, 4) possess proper physicochemical properties. These properties are crucial for determining the drug-like nature and potential practical applicability of the generated molecules, which will be elaborated on in the following paragraphs. 

\begin{table}[ht!]
    \centering
    \setlength\tabcolsep{15pt}
    \caption{The optimal ranges of 10 properties that are used to screen nearly optimal compounds, including
    seven selected ADMET properties, two physicochemical properties, and one medicinal chemistry properties. The seven ADMET properties include Caco-2 (the human colon adenocarcinoma cell lines) permeability, $\text{F}_{20\%}$ (the human oral bioavailability 20\%), Pgp-sub (the substrate of P-glycoprotein), Pgp-inh (the inhibitor of P-glycoprotein), VD (volumn density), $\text{T}_{1/2}$ (The half-life of a drug), and FDAMDD (The maximum recommended daily dose). Moreover, SAS represents the synthetic accessibility score, log $P$ is the logarithm of the n-octanol/water distribution coefficient, and log $S$ indicates the logarithm of aqueous solubility value.}
    \begin{tabular}{llll}		
        \toprule
        Property     & Profile & Excellent range  & Medium range \\ 	
        \midrule 
        Absorption   & Caco-2 permeability & $>$ -5.15 & / \\
        Absorption   & $\text{F}_{20\%}$ & 0 - 0.3  & 0.3 - 0.7  \\
        Absorption   & Pgp-sub  & 0 - 0.3  & 0.3 - 0.7  \\
        Absorption   & Pgp-inh  & 0 - 0.3  & 0.3 - 0.7  \\
        Distribution & VD & 0.04 - 20 L/kg & / \\
        Excretion    & $\text{T}_{1/2}$ &  0 - 0.3  & 0.3 - 0.7  \\
        Toxicity & FDAMDD & 0 - 0.3  & 0.3 - 0.7  \\
        Medicinal Chemistry & SAS & $<$ 6  & / \\
        Physicochemical & log $P$ & 0 - 3 log mol/L & / \\
        Physicochemical & log $S$ & -4 - 0.5 log mol/L & /\\
        \bottomrule
    \end{tabular}
    \label{tab:optimal range}
\end{table}

First, undesirable pharmacokinetics and toxicity are leading causes of drug development failure. Therefore, the assessment of absorption, distribution, metabolism, excretion, and toxicity (ADMET) properties should occur as early as possible in the drug development process. In this work, we applied \href{https://admetmesh.scbdd.com/docs/#/screening}{ADMETlab 2.0} to offer us systematic evaluation of ADMET properties, along with certain physicochemical properties and an assessment of medicinal chemistry friendliness.
In this work, we consider seven  seven ADMET properties, including Caco-2 (the human colon adenocarcinoma cell lines) permeability, $\text{F}_{20\%}$ (the human oral bioavailability 20\%), Pgp-substrate (the substrate of P-glycoprotein), Pgp-inhibitor (the inhibitor of P-glycoprotein), VD (volumn density), $\text{T}_{1/2}$ (The half-life of a drug), and FDAMDD (The maximum recommended daily dose). The optimal range of these properties are listed in \autoref{tab:optimal range}.

Second, the Lipinski's rule of five help to evaluate druglikeness or determine if a chemical compound with a certain pharmacological or biological activity has properties that would make it a likely orally active drug in humans, which should satifies four physicochemical criteria: 1) molecular weight (MW) $\le 500$ daltons, 2) octanol-water partition coefficient (log $P$) $\le 5$, 3) the number of hydrogen bond donors (nHD) $\le 5$, 4) the number of hydrogen bond acceptors (nHA) $\le 10$. 

Thirdly, synthetic accessibility is crucial to ensuring the feasibility of large-scale production of a potential drug candidate. In this study, we used RDKit to evaluate the synthetic accessibility score (SAS). A candidate drug with an SAS score less than 6 indicates that it is relatively easy to synthesize.

Lastly, physicochemical properties can significantly influence the solubility, permeability, and stability of potential drug candidates. In this study, we primarily focused on the logarithm of the n-octanol/water distribution coefficient (log $P$) and the logarithm of aqueous solubility value (log $S$). Drug candidates with a log $P$ in the range of 0 - 3 log mol/L and log $S$ in the range of -4 - 0.5 log mol/L are considered to have suitable physicochemical properties.

\section*{Code and Data availability}

The code and data are available at the public repository \href{https://github.com/wangru25/SGNC}{https://github.com/wangru25/SGNC}.

The datasets including SMILES strings and binding affinities of inhibitors targeting DAT, NET, SERT. In addition, these datasets can also be found in the 'Training Datasets' folder within the SupplementaryData.zip file, available under Supporting Information S2 for readers interested in further exploration.

Trained models from this study are saved within the aforementioned code repository. This repository includes the stochastic-based generative network complex (SGNC) developed in Python, as well as Python scripts for calculating reconstruction rates, evaluating synthetic accessibility, and generating visual plots.



\section*{Supporting Information}
The Supporting Information is available for:
\begin{itemize}
    \item[S1] Supplementary methods
    \begin{itemize}
        \item[S1.1] Fokker-Planck equation-embedded multi-target drug molecule generator
        \begin{itemize}
            \item[S1.1.1] Random variables 
            \item[S1.1.2] Wiener process and white noise 
            \item[S1.1.3] It\^{o}'s lemma 
            \item[S1.1.4] Langevin equation
            \item[S1.1.5] Derivation of Fokker-Planck equation from Langevin equation 
        \end{itemize}
        \item[S1.2] Evaluation metrics
    \end{itemize}

    \item[S2] Supplementary Data: The SupplementaryData.zip consists 3 folders, namely Training Datasets, Predictions, and Generated Molecules. 
    \begin{itemize}
        \item[S2.1] Training Datasets: This folder has the datasets used for training purposes.
        \item[S2.2] Predictions: Within this folder, one can find data related to the predicted binding affinity of inhibitors in 4 training datasets. 
        \item[S2.3] Generated Molecules: This folder documents the molecules that have been produced using the stochastic-based molecular generator.
    \end{itemize}
    \item[S3] Supplementary Figures
    \begin{itemize}
        \item[S3.1] Radar plots of physicochemical properties for 15 lead candidates
        \item[S3.2] Molecular docking and molecular interaction of 15 leads with DAT and SERT
    \end{itemize}
    \item [S4] Supplementary Dialogues
    \begin{itemize}
        \item[S4.1] The 1st persona of ChatGPT
        \item[S4.2] The 2nd persona of ChatGPT
        \item[S4.3] The 3rd persona of ChatGPT
        \item[S4.4] Other dialogues
    \end{itemize}
\end{itemize}

\section*{Competing interests}
The authors declare no competing interests.

\section*{Acknowledgment}
This work was supported in part by NIH grants R01GM126189, R01AI164266, and R35GM148196, National Science Foundation grants DMS2052983, DMS-1761320, and IIS-1900473, NASA  grant 80NSSC21M0023,   Michigan State University Research Foundation, and  Bristol-Myers Squibb  65109.


\begin{thebibliography}{10}

\bibitem{adamopoulou2020overview}
Eleni Adamopoulou and Lefteris Moussiades.
\newblock An overview of chatbot technology.
\newblock In {\em IFIP international conference on artificial intelligence
  applications and innovations}, pages 373--383. Springer, 2020.

\bibitem{bansal2018review}
Himanshu Bansal and Rizwan Khan.
\newblock A review paper on human computer interaction.
\newblock {\em Int. J. Adv. Res. Comput. Sci. Softw. Eng}, 8(4):53, 2018.

\bibitem{lyu2023translating}
Qing Lyu, Josh Tan, Mike~E Zapadka, Janardhana Ponnatapuram, Chuang Niu,
  Ge~Wang, and Christopher~T Whitlow.
\newblock Translating radiology reports into plain language using chatgpt and
  gpt-4 with prompt learning: Promising results, limitations, and potential.
\newblock {\em arXiv preprint arXiv:2303.09038}, 2023.

\bibitem{zeng2023revolutionizing}
Zehua Zeng and Hongwu Du.
\newblock Revolutionizing single cell analysis: The power of large language
  models for cell type annotation.
\newblock {\em arXiv preprint arXiv:2304.02697}, 2023.

\bibitem{white2023assessment}
Andrew~D White, Glen~M Hocky, Heta~A Gandhi, Mehrad Ansari, Sam Cox, Geemi~P
  Wellawatte, Subarna Sasmal, Ziyue Yang, Kangxin Liu, Yuvraj Singh, and
  Willmor J.~Pena Ccoa.
\newblock Assessment of chemistry knowledge in large language models that
  generate code.
\newblock {\em Digital Discovery}, 2(2):368--376, 2023.

\bibitem{wu14future}
Yijun Wu and Ailin Zhao.
\newblock Future implications of chatgpt in pharmaceutical industry: Drug
  discovery and development.
\newblock {\em Frontiers in Pharmacology}, 14:1194216.

\bibitem{savage2023drug}
Neil Savage.
\newblock Drug discovery companies are customizing chatgpt: here’s how.
\newblock {\em Nature Biotechnology}, 2023.

\bibitem{lee2023anti}
Jui-Hsuan Lee, Eric Hsiao-Kuang Wu, Yu-Yen Ou, Yueh-Che Lee, Cheng-Hsun Lee,
  and Chia-Ru Chung.
\newblock Anti-drugs chatbot: Chinese bert-based cognitive intent analysis.
\newblock {\em IEEE Transactions on Computational Social Systems}, 2023.

\bibitem{gong2022generative}
Changwei Gong, Changhong Jing, Ye~Li, Xinan Liu, Zuxin Chen, and Shuqiang Wang.
\newblock Generative artificial intelligence-enabled dynamic detection of
  nicotine-related circuits.
\newblock {\em arXiv preprint arXiv:2212.06330}, 2022.

\bibitem{feng2022machine}
Hongsong Feng, Kaifu Gao, Dong Chen, Li~Shen, Alfred~J Robison, Edmund
  Ellsworth, and Guo-Wei Wei.
\newblock Machine learning analysis of cocaine addiction informed by dat, sert,
  and net-based interactome networks.
\newblock {\em Journal of chemical theory and computation}, 18(4):2703--2719,
  2022.

\bibitem{yang2023deep}
Yuwei Yang, Chang-Yu Hsieh, Yu~Kang, Tingjun Hou, Huanxiang Liu, and Xiaojun
  Yao.
\newblock Deep generation model guided by the docking score for active
  molecular design.
\newblock {\em Journal of Chemical Information and Modeling}, 2023.

\bibitem{pan2022aa}
Xiaolin Pan, Hao Wang, Yueqing Zhang, Xingyu Wang, Cuiyu Li, Changge Ji, and
  John~ZH Zhang.
\newblock Aa-score: a new scoring function based on amino acid-specific
  interaction for molecular docking.
\newblock {\em Journal of Chemical Information and Modeling},
  62(10):2499--2509, 2022.

\bibitem{ballester2010machine}
Pedro~J Ballester and John~BO Mitchell.
\newblock A machine learning approach to predicting protein--ligand binding
  affinity with applications to molecular docking.
\newblock {\em Bioinformatics}, 26(9):1169--1175, 2010.

\bibitem{gao2020generative}
Kaifu Gao, Duc~Duy Nguyen, Meihua Tu, and Guo-Wei Wei.
\newblock Generative network complex for the automated generation of drug-like
  molecules.
\newblock {\em Journal of chemical information and modeling},
  60(12):5682--5698, 2020.

\bibitem{grow2019generative}
Christopher Grow, Kaifu Gao, Duc~Duy Nguyen, and Guo-Wei Wei.
\newblock Generative network complex (gnc) for drug discovery.
\newblock {\em Communications in information and systems}, 19(3):241, 2019.

\bibitem{gaulton2012chembl}
Anna Gaulton, Louisa~J Bellis, A~Patricia Bento, Jon Chambers, Mark Davies,
  Anne Hersey, Yvonne Light, Shaun McGlinchey, David Michalovich, Bissan
  Al-Lazikani, and John~P. Overington.
\newblock Chembl: a large-scale bioactivity database for drug discovery.
\newblock {\em Nucleic acids research}, 40(D1):D1100--D1107, 2012.

\bibitem{gao2021proteome}
Kaifu Gao, Dong Chen, Alfred~J Robison, and Guo-Wei Wei.
\newblock Proteome-informed machine learning studies of cocaine addiction.
\newblock {\em The journal of physical chemistry letters}, 12(45):11122--11134,
  2021.

\bibitem{gupta2021machine}
Aayush Gupta and Huan-Xiang Zhou.
\newblock Machine learning-enabled pipeline for large-scale virtual drug
  screening.
\newblock {\em Journal of Chemical Information and Modeling}, 61(9):4236--4244,
  2021.

\bibitem{xu2022geodiff}
Minkai Xu, Lantao Yu, Yang Song, Chence Shi, Stefano Ermon, and Jian Tang.
\newblock Geodiff: A geometric diffusion model for molecular conformation
  generation.
\newblock {\em arXiv preprint arXiv:2203.02923}, 2022.

\bibitem{risken1996fokker}
Hannes Risken and Hannes Risken.
\newblock {\em Fokker-planck equation}.
\newblock Springer, 1996.

\bibitem{ozturk2016comparative}
Hakime {\"O}zt{\"u}rk, Elif Ozkirimli, and Arzucan {\"O}zg{\"u}r.
\newblock A comparative study of smiles-based compound similarity functions for
  drug-target interaction prediction.
\newblock {\em BMC bioinformatics}, 17(1):1--11, 2016.

\bibitem{yang2013molecular}
Jane~Y. Yang, Laura~M. Sanchez, Christopher~M. Rath, Xueting Liu, Paul~D.
  Boudreau, Nicole Bruns, Evgenia Glukhov, Anne Wodtke, Rafael de~Felicio,
  Amanda Fenner, Weng~Ruh Wong, Roger~G. Linington, Lixin Zhang, Hosana~M.
  Debonsi, William~H. Gerwick, and Pieter~C. Dorrestein.
\newblock Molecular networking as a dereplication strategy.
\newblock {\em Journal of natural products}, 76(9):1686--1699, 2013.

\bibitem{chen2020analysis}
Xiaoxia Chen, Hao Li, Lichao Tian, Qinwei Li, Jinxiang Luo, and Yongqiang
  Zhang.
\newblock Analysis of the physicochemical properties of acaricides based on
  lipinski's rule of five.
\newblock {\em Journal of computational biology}, 27(9):1397--1406, 2020.

\bibitem{flower2002drug}
Darren~R Flower.
\newblock {\em Drug design: cutting edge approaches}, volume 279.
\newblock Royal Society of Chemistry, 2002.

\bibitem{probst2018smilesdrawer}
Daniel Probst and Jean-Louis Reymond.
\newblock Smilesdrawer: parsing and drawing smiles-encoded molecular structures
  using client-side javascript.
\newblock {\em Journal of chemical information and modeling}, 58(1):1--7, 2018.

\bibitem{trott2010autodock}
Oleg Trott and Arthur~J Olson.
\newblock Autodock vina: improving the speed and accuracy of docking with a new
  scoring function, efficient optimization, and multithreading.
\newblock {\em Journal of computational chemistry}, 31(2):455--461, 2010.

\bibitem{cang2018integration}
Zixuan Cang and Guo-Wei Wei.
\newblock Integration of element specific persistent homology and machine
  learning for protein-ligand binding affinity prediction.
\newblock {\em International journal for numerical methods in biomedical
  engineering}, 34(2):e2914, 2018.

\bibitem{kalliokoski2013comparability}
Tuomo Kalliokoski, Christian Kramer, Anna Vulpetti, and Peter Gedeck.
\newblock Comparability of mixed ic50 data--a statistical analysis.
\newblock {\em PloS one}, 8(4):e61007, 2013.

\bibitem{winter2019learning}
Robin Winter, Floriane Montanari, Frank No{\'e}, and Djork-Arn{\'e} Clevert.
\newblock Learning continuous and data-driven molecular descriptors by
  translating equivalent chemical representations.
\newblock {\em Chemical science}, 10(6):1692--1701, 2019.

\bibitem{landrum2013rdkit}
Greg Landrum.
\newblock Rdkit: A software suite for cheminformatics, computational chemistry,
  and predictive modeling.
\newblock {\em Greg Landrum}, 2013.

\bibitem{corso2022diffdock}
Gabriele Corso, Hannes St{\"a}rk, Bowen Jing, Regina Barzilay, and Tommi
  Jaakkola.
\newblock Diffdock: Diffusion steps, twists, and turns for molecular docking.
\newblock {\em arXiv preprint arXiv:2210.01776}, 2022.

\end{thebibliography}

\end{document}


\title{Chatbots in Drug Discovery: A Case Study on Anti-Cocaine Addiction Drug Development with ChatGPT}
\author{Rui Wang$^1$, Hongsong Feng$^1$, and Guo-Wei Wei$^{1,2,3}$\footnote{
		Corresponding author.		Email: weig@msu.edu} \\
$^1$ Department of Mathematics, \\
Michigan State University, MI 48824, USA.\\
$^2$ Department of Electrical and Computer Engineering,\\
Michigan State University, MI 48824, USA. \\
$^3$ Department of Biochemistry and Molecular Biology,\\
Michigan State University, MI 48824, USA. \\
}
 
\date{\today} 

\maketitle

{\setcounter{tocdepth}{4} \tableofcontents}
\clearpage \pagebreak \setcounter{page}{1}
\renewcommand{\thepage}{{\arabic{page}}}

\section{Supplementary methods}

\subsection{Derivation of Fokker-Planck equation from Langevin equation}
For an It\^{o} process driven by the standard Wiener process $W_t$ in one spatial dimension $x$, a stochastic differential equation (SDE) is given in the form
            \begin{equation}\label{eq:sde}
                dX_t = \mu(X_t, t) dt + \sigma(X_t, t) dW_t
            \end{equation}
            where $\mu_t = \mu(X_t, t)$ and $\sigma_t = \sigma(X_t, t)$ are given functions representing the drift and diffusion coefficients, respectively. To derive the Fokker-Planck equation (also called forward kolmogorov equation), we can first apply It\^{o}' lemma. Assume $f(x)$ is a twice-differential function, then its expansion in Taylor's series is:
            \[
            df = f_x dX_t + \frac{1}{2} f_{xx} dX_t^2 + \dots.
            \]
            Substituting the Eq. \eqref{eq:sde} into this, we have:
            \begin{align}\label{eq:Ito Lemma}
                df & = f_x(\mu_t dt + \sigma_tdW_t) + \frac{1}{2} f_{xx} (\mu_t dt + \sigma_t W_t)^2 + ...\\
                   & = f_x\mu_t dt + f_x \sigma_tdW_t + \frac{1}{2}f_{xx} (\mu_t^2 dt^2 + 2\mu_t\sigma_tdtdW_t + \sigma_t^2dW_t^2).
            \end{align}
            In the limit $dt \to 0$, the terms $dt^2$ and $dt Wt$ tend to zero faster than $dW_t^2$. Meanwhile, since $W_t$ is a Wiener process, which assures $dW_t^2 = dt$. Therefore, from Eq. \eqref{eq:Ito Lemma}, we can obtain 
            \[
            df = (\mu_tf_x + \frac{1}{2}f_xx\sigma_t^2)dt + \sigma_tf_xdW_t,
            \]
            then its expected value would be:
            \begin{align*}
                E(df) & = E[(\mu_tf_x + \frac{1}{2}f_{xx}\sigma_t^2)dt + \sigma_tf_xdW_t]\\
                      & = E[(\mu_tf_x + \frac{1}{2}f_{xx}\sigma_t^2)dt] + E[\sigma_tf_xdW_t]\\
                      & = E[(\mu_tf_x + \frac{1}{2}f_{xx}\sigma_t^2)dt] + 0.\\
            \end{align*}
            Rearranging terms, we can get:
            \[
            \frac{dE[f]}{dt} = E[\mu_tf_x + \frac{1}{2}f_{xx}\sigma_t^2].
            \]
            Let $p(x,t)$ to be the probability density function of the random variable $x$. Then by applying the definition of expected value, we have:
            \begin{align*}
                \frac{d}{dt}\int_{-\infty}^{+\infty} f(x)p(x,t)dx 
                & = \int_{-\infty}^{+\infty}(\mu_tf_x  + \frac{1}{2}f_{xx}\sigma_t^2)p(x,t) dx\\
                & = \int_{-\infty}^{+\infty}\mu_tf_xp(x,t)dx + \frac{1}{2}\int_{-\infty}^{+\infty}f_{xx}\sigma_t^2)p(x,t) dx\\
                & = -\int_{-\infty}^{+\infty}f(x)\mu_t\frac{\partial p}{\partial x}dx - \frac{1}{2}\int_{-\infty}^{+\infty}\sigma_tf_x\frac{\partial P}{\partial x}dx\\
                & = -\int_{-\infty}^{+\infty}f(x)\mu_t\frac{\partial p}{\partial x}dx + \frac{1}{2}\int_{-\infty}^{+\infty}f(x)\sigma_t^2\frac{\partial^2 p}{\partial x^2}dx\\
                & = \int_{-\infty}^{+\infty}f(x)(-\mu_t\frac{\partial p}{\partial x} + \frac{1}{2}\sigma_t^2\frac{\partial^2 p}{\partial x^2})dx.
            \end{align*}
            Therefore, we can get the Fokker-Planck Equation of the probability density function $p(x,t)$ of the random variable $X_t$ in the one spatial dimension to be:
            \begin{equation}
                \frac{\partial p}{\partial t} = -\mu_t\frac{\partial p}{\partial x} + \frac{1}{2}\sigma_t^2\frac{\partial^2 p}{\partial x^2}
            \end{equation}

\subsection{Evaluation metrics}
To evaluate our binding affinity prediction models, we utilized two metrics: the Pearson Correlation Coefficient (R) and the Root Mean Squared Error (RMSE). These metrics are defined as follows:

Given two vectors ${\bf{x}}=(x_1,x_2,\cdots,x_n)$ and ${\bf{y}}=(y_1,y_2,\cdots,y_n)$, the PCC between these vectors is:

\begin{align}
\text{R}=\frac{\sum(x_i-\bar{{\bf{x}}})(y_i-\bar{{\bf{y}}})}{\sqrt{\sum (x_i-\bar{{\bf{x}}})^2\sum (y_i-\bar{{\bf{y}}})^2}},
\end{align}
where $\bar{{\bf{x}}}$ and $\bar{{\bf{y}}}$ denote the means of vectors ${\bf{x}}$ and ${\bf{y}}$, respectively. Also, the RMSE is defined as:
\begin{align}
\text{RMSE} = \sqrt{\frac{1}{n}\sum_{i=1}^{n}(y_i-\hat{y}_i)^2},
\end{align}
where $y_i$ and $\hat{y}_i$ are the true and predicted values of the $i$th sample.

Besides, the similarities between generated molecules and four inhibitor datasets are also analyzed. Given two $n$-dimensional vectors $\mathbf{v}_1$ and $\mathbf{v}_2$, the cosine similarity $S_{C}(\mathbf{v}_1,\mathbf{v}_2)$ is defined as
\begin{equation}
    S_{C}(\mathbf{v}_1,\mathbf{v}_2)=\frac{\mathbf {v_1} \cdot \mathbf {v_2}}{\|\mathbf {v_1} \|\|\mathbf {v_2} \|}
\end{equation}

\section{Supplementary Data}
The file SupplementaryData.zip comprises three folders:
\begin{enumerate}
    \item Training Datasets: This folder has the datasets used for training purposes.
    \item Predictions: Within this folder, one can find data related to the predicted binding affinity of inhibitors in 4 training datasets. 
    \item Generated Molecules: This folder documents the molecules that have been produced using the stochastic-based molecular generator. 
\end{enumerate}

\subsection{Training Datasets} 
This folder contains 4 datasets, each employed to train separate binding affinity predictors. The datasets are labeled as DAT-Inhibitor, NET-Inhibitors, SERT-Inhibitors, and hERG-Inhibitors. Within each dataset, there are SMILES strings representing different inhibitors, paired with their respective labels, which indicates the experimental binding affinities.

\subsection{Predictions}
This folder saves the predicted binding affinity of each inhibitors in our 4 training datasets, including DAT\_properties.csv, NET\_properties.csv, SERT\_properties.csv, extendex\_hERG\_properties.csv

\subsection{Generated Molecules}
Generated Molecules Folder contains 4 csv files and 1 folder, including
\begin{enumerate}
    \item In the Generated Molecules Folder, folder 16M\_molecules\_properties contains 1) the predicted binding affinities of generated molecules to DAT, SERT, NET, and hERG, and 2) the similarities of generated molecules to Reference 1 from DAT-Inhibitors dataset, Reference 2 from NET-Inhibitors dataset, and Reference 3 from SERT-Inhibitors dataset. 
    \item File 330\_molecules\_w\_proper\_BA.csv catalogs the ADMET properties of 330 generated molecules. These molecules have binding affinities ($\Delta G$) less than -9.54 kcal/mol for DAT, NET, and SERT, and greater than -8.18 kcal/mol for hERG.
    \item File 15\_candidate\_leads.csv contains a selection of molecules whose ADMET properties lie within the medium range. 
    \item File 2nd\_round\_generated\_molecules\_input.csv records the SMILES strings of generated molecules. This is after the removal of duplicates and molecules that could not be parsed by RDKit.
    \item The SMILES strings of these generated molecules are used as input for the encoder of the SeqSeq AE. The resulting decoded SMILES strings have been saved in the 2nd\_round\_generated\_molecules\_input.csv file
\end{enumerate}

\section{Supplenmentary Figures}
\subsection{Radar plots of physicochemical properties for 15 lead candidates}

\begin{figure}[ht!]
    \includegraphics[width=1.0\textwidth]{RadarPlot_SI.pdf}
    \centering
    \caption{Physicochemical properties of 15 lead candidates, including MW (molecular weight), log P (logarithm of octanol/water partition coefficient), log S (logarithm of the aqueous solubility), log D (log P at physiological pH 7.4), nHA (number of hydrogen bond acceptors), nHD (number of hydrogen bond donors), TPSA (topological polar surface area), nRot (number of rotatable bonds), nRing (number of rings), MaxRing (number of atoms in the largest ring), nHet (number of heteroatoms), fChar (formal charge), and nRig (number of rigid bonds). Here the purple dots denote the minimal value and the blue dots indicate the maximal value within the optimal range. The red lines represent the values of the properties for each lead candidate.}
    \label{fig:combine 2}
\end{figure}

\subsection{Molecular docking and molecular interaction of 15 leads with DAT and SERT}

\begin{figure}[htbp!]
    \includegraphics[width=1.0\textwidth]{DAT_Docking_SI.pdf}
    \centering
    \caption{3D structure of molecular docking for candidate leads with DAT.}
    \label{fig:DAT docking}
\end{figure}

\begin{figure}[htbp!]
    \includegraphics[width=1.0\textwidth]{DAT_Ligplot_SI.pdf}
    \centering
    \caption{Molecular interaction of 15 leads with DAT.}
    \label{fig:DAT Ligplot}
\end{figure}

\begin{figure}[htbp!]
    \includegraphics[width=1.0\textwidth]{SERT_Docking_SI.pdf}
    \centering
    \caption{3D structure of molecular docking for candidate leads with SERT.}
    \label{fig:DAT docking}
\end{figure}

\begin{figure}[htbp!]
    \includegraphics[width=1.0\textwidth]{SERT_Ligplot_SI.pdf}
    \centering
    \caption{Molecular interaction of 15 leads with SERT.}
    \label{fig:SERT Ligplot}
\end{figure}



\title{Chatbots in Drug Discovery: A Case Study on Anti-Cocaine Addiction Drug Development with ChatGPT}
\author{Rui Wang$^1$, Hongsong Feng$^1$, and Guo-Wei Wei$^{1,2,3}$\footnote{
		Corresponding author.		Email: weig@msu.edu} \\
$^1$ Department of Mathematics, \\
Michigan State University, MI 48824, USA.\\
$^2$ Department of Electrical and Computer Engineering,\\
Michigan State University, MI 48824, USA. \\
$^3$ Department of Biochemistry and Molecular Biology,\\
Michigan State University, MI 48824, USA. \\
}
 
\date{\today} 

\maketitle

{\setcounter{tocdepth}{4} \tableofcontents}
\clearpage \pagebreak \setcounter{page}{1}
\renewcommand{\thepage}{{\arabic{page}}}

\section{Supplementary Methods}

\subsection{Fokker-Planck equation-embedded multi-target drug molecule generator}
\subsubsection{Random variables}
The Fokker-Planck equation describes the time evolution of a distribution function driven by a forcing term and a random term \cite{risken1996fokker}. 
A random variable $X$ is a variable whose possible values are outcomes of a random phenomenon. The random variable can be either discrete, taking on a countable number of values, or continuous, taking on any value within a certain range or set. For a discrete random variable, we can write $P(X=x)$ to denote the probability that $X$ takes the value $x$. For a continuous random variable, we talk about probability density function (pdf) $f_X(x)$ such that for any interval $[a, b]$, $ P(a \leq X \leq b) = \int_{a}^{b} f_X(x) \, dx$.

The notation $\langle R(n)R(m) \rangle$ typically denotes the expected value of the product of two random variables $R(n)$ and $R(m)$. This is often encountered in the context of statistical physics and signal processing, where it is referred to as the autocorrelation function. In details, $R(n)$ and $R(m)$ are random variables, often representing some quantity of interest at different times or positions $n$ and $m$, respectively.
The product $R(n)R(m)$ represents the joint behavior of these two quantities.
The angle brackets $\langle \cdot \rangle$ denote the expectation value or mean value, which is a fundamental concept in probability theory and statistics.
So, $\langle R(n)R(m) \rangle$ is the average value of the product $R(n)R(m)$, which gives information about how $R$ at time/position $n$ is correlated with $R$ at time/position $m$.

If $R(n)$ represents a stochastic or random process (such as a fluctuating quantity in time or space), then $\langle R(n)R(m) \rangle$ gives the autocorrelation function of the process, which measures how much the process at one point in time or space is correlated with the process at another point. For example, if $R(n)$ represents the position of a particle undergoing random motion at time $n$, then $\langle R(n)R(m) \rangle$ might represent the average over many realizations of the motion, which gives statistical information about how the position of the particle at one time is related to its position at another time.

\subsubsection{Wiener process and white noise}
{\bf A Wiener process} (also known as Brownian motion) is a continuous-time stochastic process that is widely used in physics and finance. It is characterized by the following properties:
\begin{enumerate}
    \item[1)] The process starts at zero: $W(0) = 0$.
    \item[2)] The increments are independent: for any $0 \leq s < t$, the increment $W(t) - W(s)$ is independent of the history of the process up to time $s$.
    \item[3)] The increments are normally distributed: for any $0 \leq s < t$, the increment $W(t) - W(s)$ follows a normal distribution with mean 0 and variance $t-s$.
    \item[4)] The paths are continuous: although the process is nowhere differentiable.
\end{enumerate}

{\bf White noise} is a mathematical idealization of a random signal that is used in many fields, from acoustics to physics to statistics. In the context of stochastic processes, a white noise process is a sequence of uncorrelated random variables that are identically distributed. 

Properties of white noise include:
\begin{enumerate}
    \item[1)]  The expected value is zero: $E[\xi(t)] = 0$.
    \item[2)] The autocorrelation function is a delta function, which means the variables are uncorrelated: $E[\xi(t)\xi(t+\tau)] = \sigma^2\delta(\tau)$, where $\delta(\tau)$ is the Dirac delta function.
    \item[3)] The power spectral density (the Fourier transform of the autocorrelation function) is constant, which means the signal contains all frequencies with equal intensity. This is the property that gives white noise its name, by analogy with white light.
\end{enumerate}

It's worth noting that in continuous time, the concept of white noise is more difficult to define rigorously, as it essentially involves taking the limit of a sequence of independent random variables in such a way that each individual variable has infinitesimal variance, but the total variance remains finite. This is often formalized in terms of the concept of a "generalized random process" or "distribution-valued process". The derivative of a Wiener process is often taken as a model of white noise in continuous time.

\subsubsection{Ito's lemma}
Ito's lemma is a fundamental result in stochastic calculus that provides a rule for finding the differential of a function of a stochastic process. Named after Kiyoshi Ito, a Japanese mathematician who made significant contributions to the theory of stochastic processes, Ito's lemma is a key tool for manipulating stochastic differential equations (SDEs). It\^{o}'s lemma can be thought of as a version of the chain rule for stochastic calculus, but unlike the chain rule, it includes an additional term due to the stochastic nature of the differential. Suppose we have a stochastic process $X_t$ that satisfies the SDE:

\begin{equation}
dX_t = \mu dt + \sigma dW_t
\end{equation}

where $W_t$ is a Wiener process, and $\mu$ and $\sigma$ are, in the simplest case, constants but can also be functions of $t$ and $X_t$. Suppose we also have a twice-differentiable function $f(t, X_t)$.

Ito's lemma states that the differential $df(t, X_t)$ is given by:

\begin{equation}
df(t, X_t) = \frac{\partial f}{\partial t} dt + \frac{\partial f}{\partial x} dX_t + \frac{1}{2} \frac{\partial^2 f}{\partial x^2} (dX_t)^2
\end{equation}

Now, because $(dX_t)^2 = (\mu dt + \sigma dW_t)^2$ and using the property that $(dW_t)^2 = dt$, the second order term becomes $\sigma^2 dt$. So the differential $df$ can be expressed as:

\begin{equation}
df(t, X_t) = \left( \frac{\partial f}{\partial t} + \mu \frac{\partial f}{\partial x} + \frac{1}{2} \sigma^2 \frac{\partial^2 f}{\partial x^2} \right) dt + \sigma \frac{\partial f}{\partial x} dW_t
\end{equation}

The additional second derivative term (the last term in the parentheses on the right-hand side) is due to the stochastic nature of the differential, and is often referred to as the Ito correction term. This term is what makes Ito calculus distinct from ordinary calculus.

\subsubsection{Langevin equation}\label{subsubsec:Langevin equation}
The Langevin equation is a stochastic differential equation (SDE) that describes the time evolution of a system subjected to both deterministic forces and random forces. It is named after the French physicist Paul Langevin, who introduced the equation in 1908 to describe the motion of a particle undergoing Brownian motion.

The Langevin equation is used to model various physical systems, including particles in a fluid, charged particles in electromagnetic fields, and particles in systems with thermal noise. It is particularly useful for studying systems where fluctuations play a significant role in the system's behavior.

The Langevin equation is a commonly used stochastic differential equation (SDE) in physical that aims to describe the behavior of a system as it evolves over time under the influence of deterministic and random (fluctuating) forces. Langevin equation can describe the motion of a particle in a fluid,
\begin{equation}
    m\frac{d\mathbf{v}}{dt} = -\lambda \mathbf{v} + \mathbf{\eta}(t),
\end{equation}
where $m$ is the mass of the particle, $\mathbf{v}$ is the velocity of the particle, $\lambda$ is its corresponding damping coefficient, and $\mathbf{\eta}$ is the noise term which represent the effect of the collisions with the molecules of the fluid. In many cases, the 1-dimensional Langevin equation is simplified and written in a general form as:
\begin{equation}
    \frac{dx}{dt} = - \gamma x + \xi(t),
\end{equation}
where $\xi(t)$ is a Gaussian white noise process with $\langle \xi(t) \rangle = 0$ and $\langle \xi(t) \xi(t') \rangle = \delta(t - t')$. The general solution of 1-dimensional Langevin equation has the form:
\begin{equation}\label{eq:1d Langevin eq sol}
    x(t) = Ce^{-\gamma t} + \int_0^t e^{-\gamma(t-u)}\xi(u)du,
\end{equation}
where the initial state $x(0) = C$.

In this work, we aim to employ Langevin equation to better design the molecule. We assume $\mathbf{X}$ is a latent space vector of a molecule with 512 dimensions, and $\mathbf{X}_k$ represents its $k$-th latent space reference vector. The the Langevin equation of this system is:
\begin{equation}
    \frac{d\mathbf{X}}{dt} = \alpha \sum_k a_k(\mathbf{X}_k - \mathbf{X}) + \bm{\xi}(t),
\end{equation}
where $a_k$ is a positive weighting parameter corresponds to $\mathbf{X}_k$ statisfying $\displaystyle \sum_k a_k =1$. Then according to Eq. \eqref{eq:1d Langevin eq sol}, the general solution of this system is given by:
\begin{equation}\label{eq:1d Langevin eq}
    \mathbf{X}(t) = \mathbf{C}^{-\alpha t} + \int_0^t e^{-\alpha(t-u)} (\alpha\sum_k a_k\mathbf{X}_k + \bm{\xi}(u))du,
\end{equation}
where the initial state $\mathbf{X}(0) = \mathbf{C}$.

\subsubsection{Derivation of Fokker-Planck equation from Langevin equation}
The Fokker–Planck equation is a kinetic equation  describing  the time evolution of the probability density distribution under the influence of drag and random forces \cite{risken1996fokker}. It can be derived in different ways. 
 For an It\^{o} process driven by the standard Wiener process $W_t$ in one spatial dimension $x$, a stochastic differential equation (SDE) is given in the form
            \begin{equation}\label{eq:sde}
                dX_t = \mu(X_t, t) dt + \sigma(X_t, t) dW_t
            \end{equation}
            where $\mu_t = \mu(X_t, t)$ and $\sigma_t = \sigma(X_t, t)$ are given functions representing the drift and diffusion coefficients, respectively. To derive the Fokker-Planck equation (also called forward kolmogorov equation), we can first apply It\^{o}' lemma. Assume $f(x)$ is a twice-differential function, then its expansion in Taylor's series is:
            \[
            df = f_x dX_t + \frac{1}{2} f_{xx} dX_t^2 + \dots.
            \]
            Substituting the Eq. \eqref{eq:sde} into this, we have:
            \begin{align}\label{eq:Ito Lemma}
                df & = f_x(\mu_t dt + \sigma_tdW_t) + \frac{1}{2} f_{xx} (\mu_t dt + \sigma_t W_t)^2 + ...\\
                   & = f_x\mu_t dt + f_x \sigma_tdW_t + \frac{1}{2}f_{xx} (\mu_t^2 dt^2 + 2\mu_t\sigma_tdtdW_t + \sigma_t^2dW_t^2).
            \end{align}
            In the limit $dt \to 0$, the terms $dt^2$ and $dt Wt$ tend to zero faster than $dW_t^2$. Meanwhile, since $W_t$ is a Wiener process, which assures $dW_t^2 = dt$. Therefore, from Eq. \eqref{eq:Ito Lemma}, we can obtain 
            \[
            df = (\mu_tf_x + \frac{1}{2}f_xx\sigma_t^2)dt + \sigma_tf_xdW_t,
            \]
            then its expected value would be:
            \begin{align*}
                E(df) & = E[(\mu_tf_x + \frac{1}{2}f_{xx}\sigma_t^2)dt + \sigma_tf_xdW_t]\\
                      & = E[(\mu_tf_x + \frac{1}{2}f_{xx}\sigma_t^2)dt] + E[\sigma_tf_xdW_t]\\
                      & = E[(\mu_tf_x + \frac{1}{2}f_{xx}\sigma_t^2)dt] + 0.\\
            \end{align*}
            Rearranging terms, we can get:
            \[
            \frac{dE[f]}{dt} = E[\mu_tf_x + \frac{1}{2}f_{xx}\sigma_t^2].
            \]
            Let $p(x,t)$ to be the probability density function of the random variable $x$. Then by applying the definition of expected value, we have:
            \begin{align*}
                \frac{d}{dt}\int_{-\infty}^{+\infty} f(x)p(x,t)dx 
                & = \int_{-\infty}^{+\infty}(\mu_tf_x  + \frac{1}{2}f_{xx}\sigma_t^2)p(x,t) dx\\
                & = \int_{-\infty}^{+\infty}\mu_tf_xp(x,t)dx + \frac{1}{2}\int_{-\infty}^{+\infty}f_{xx}\sigma_t^2)p(x,t) dx\\
                & = -\int_{-\infty}^{+\infty}f(x)\mu_t\frac{\partial p}{\partial x}dx - \frac{1}{2}\int_{-\infty}^{+\infty}\sigma_tf_x\frac{\partial P}{\partial x}dx\\
                & = -\int_{-\infty}^{+\infty}f(x)\mu_t\frac{\partial p}{\partial x}dx + \frac{1}{2}\int_{-\infty}^{+\infty}f(x)\sigma_t^2\frac{\partial^2 p}{\partial x^2}dx\\
                & = \int_{-\infty}^{+\infty}f(x)(-\mu_t\frac{\partial p}{\partial x} + \frac{1}{2}\sigma_t^2\frac{\partial^2 p}{\partial x^2})dx.
            \end{align*}
            Therefore, we can get the Fokker-Planck Equation of the probability density function $p(x,t)$ of the random variable $X_t$ in the one spatial dimension to be:
            \begin{equation}
                \frac{\partial p}{\partial t} = -\mu_t\frac{\partial p}{\partial x} + \frac{1}{2}\sigma_t^2\frac{\partial^2 p}{\partial x^2}
            \end{equation}

\subsection{Evaluation metrics}
To evaluate our binding affinity prediction models, we utilized two metrics: the Pearson Correlation Coefficient (R) and the Root Mean Squared Error (RMSE). These metrics are defined as follows:

Given two vectors ${\bf{x}}=(x_1,x_2,\cdots,x_n)$ and ${\bf{y}}=(y_1,y_2,\cdots,y_n)$, the PCC between these vectors is:

\begin{align}
\text{R}=\frac{\sum(x_i-\bar{{\bf{x}}})(y_i-\bar{{\bf{y}}})}{\sqrt{\sum (x_i-\bar{{\bf{x}}})^2\sum (y_i-\bar{{\bf{y}}})^2}},
\end{align}
where $\bar{{\bf{x}}}$ and $\bar{{\bf{y}}}$ denote the means of vectors ${\bf{x}}$ and ${\bf{y}}$, respectively. Also, the RMSE is defined as:
\begin{align}
\text{RMSE} = \sqrt{\frac{1}{n}\sum_{i=1}^{n}(y_i-\hat{y}_i)^2},
\end{align}
where $y_i$ and $\hat{y}_i$ are the true and predicted values of the $i$th sample.

Besides, the similarities between generated molecules and four inhibitor datasets are also analyzed. Given two $n$-dimensional vectors $\mathbf{v}_1$ and $\mathbf{v}_2$, the cosine similarity $S_{C}(\mathbf{v}_1,\mathbf{v}_2)$ is defined as
\begin{equation}
    S_{C}(\mathbf{v}_1,\mathbf{v}_2)=\frac{\mathbf {v_1} \cdot \mathbf {v_2}}{\|\mathbf {v_1} \|\|\mathbf {v_2} \|}
\end{equation}

\section{Supplementary Data}
The file SupplementaryData.zip comprises three folders:
\begin{enumerate}
    \item Training Datasets: This folder has the datasets used for training purposes.
    \item Predictions: Within this folder, one can find data related to the predicted binding affinity of inhibitors in 4 training datasets. 
    \item Generated Molecules: This folder documents the molecules that have been produced using the stochastic-based molecular generator. 
\end{enumerate}

\subsection{Training Datasets} 
This folder contains 4 datasets, each employed to train separate binding affinity predictors. The datasets are labeled as DAT-Inhibitor, NET-Inhibitors, SERT-Inhibitors, and hERG-Inhibitors. Within each dataset, there are SMILES strings representing different inhibitors, paired with their respective labels, which indicates the experimental binding affinities.

\subsection{Predictions}
This folder saves the predicted binding affinity of each inhibitors in our 4 training datasets, including DAT\_properties.csv, NET\_properties.csv, SERT\_properties.csv, extendex\_hERG\_properties.csv

\subsection{Generated Molecules}
Generated Molecules Folder contains 4 csv files and 1 folder, including
\begin{enumerate}
    \item In the Generated Molecules Folder, folder 16M\_molecules\_properties contains 1) the predicted binding affinities of generated molecules to DAT, SERT, NET, and hERG, and 2) the similarities of generated molecules to Reference 1 from DAT-Inhibitors dataset, Reference 2 from NET-Inhibitors dataset, and Reference 3 from SERT-Inhibitors dataset. 
    \item File 330\_molecules\_w\_proper\_BA.csv catalogs the ADMET properties of 330 generated molecules. These molecules have binding affinities ($\Delta G$) less than -9.54 kcal/mol for DAT, NET, and SERT, and greater than -8.18 kcal/mol for hERG.
    \item File 15\_candidate\_leads.csv contains a selection of molecules whose ADMET properties lie within the medium range. 
    \item File 2nd\_round\_generated\_molecules\_input.csv records the SMILES strings of generated molecules. This is after the removal of duplicates and molecules that could not be parsed by RDKit.
    \item The SMILES strings of these generated molecules are used as input for the encoder of the SeqSeq AE. The resulting decoded SMILES strings have been saved in the 2nd\_round\_generated\_molecules\_input.csv file
\end{enumerate}

\section{Supplementary Figures}

\subsection{Radar plots of physicochemical properties for 15 lead candidates}

In this study, we utilized generative network complex to design multi-target effective compounds that can have potential for the treatment of anti-cocaine addiction. To identify those drug-like compounds among the generated molecules, we considered ADMET properties, physicochemical properties, and medicinal chemistry properties. Based on seven selected ADMET properties, two physicochemical properties, and one medicinal chemistry property, we conducted a meticulous screening, successfully pinpointing 15 nearly-optimal lead candidates. There are many other important physicochemical properties that need to be evaluated in designing a drug. We leveraged the ADMETlab2 server \cite{xiong2021admetlab} to predict additional physicochemical properties for these 15 drug candidates, incorporating a total of 13 physicochemical properties in our analysis. Table \ref{tab:property-optimal} indicates the optimal ranges of the 13 additional physicochemical properties. 

Figure \ref{fig:combine 2} shows the additional physicochemical properties of these 15 compounds, predicted by ADMETlab2 server. The purple dots denote the minimal value and the blue dots indicate the maximal value within the optimal range. The red lines represent the values of the properties for each compound. It is observed that all these compounds have their physicochemical properties in the optimal ranges. 

\begin{table}
	\centering
	\begin{tabular}{c|c |c }		
		\hline
		Property Names& Formulations of properties & Optimal range   \\ 	
		\hline
		MW & Molecular Weight, contains hydrogen atoms &100--600\\
		nHA	 & Number of hydrogen bond acceptors & 0--12\\
		nHD	 & Number of hydrogen bond donor & 0--7\\
		nRot & Number of rotatable bonds & 0--11  \\
		nRing & Number of rings &0--6 \\
		MaxRing & Number of atoms in the biggest ring & 0--18 \\
		nHet & Number of heteroatoms &1--15		\\
		fChar	& Formal charge & -4-4	\\
		nRig &	Number of rigid bonds & 0--30	\\
		TPSA& Topological polar surface area &	0--140	\\
		logS& log of the aqueous  solubility &	-4--0.5 log mol/L	\\
		logP& log of octanol/water partition coefficient &	0--3		\\
		logD & logP at physiological pH 7.4&	1--3 \\
		\hline
	\end{tabular}
	\caption{The optimal ranges of  13 selected ADMET characteristics.}
	\label{tab:property-optimal}
\end{table}

\begin{figure}[ht!]
    \includegraphics[width=1.0\textwidth]{RadarPlot_SI.pdf}
    \centering
    \caption{Physicochemical properties of 15 lead candidates, including MW (molecular weight), log P (logarithm of octanol/water partition coefficient), log S (logarithm of the aqueous solubility), log D (log P at physiological pH 7.4), nHA (number of hydrogen bond acceptors), nHD (number of hydrogen bond donors), TPSA (topological polar surface area), nRot (number of rotatable bonds), nRing (number of rings), MaxRing (number of atoms in the largest ring), nHet (number of heteroatoms), fChar (formal charge), and nRig (number of rigid bonds). Here the purple dots denote the minimal value and the blue dots indicate the maximal value within the optimal range. The red lines represent the values of the properties for each lead candidate.}
    \label{fig:combine 2}
\end{figure}

\subsection{Molecular docking and molecular interaction of 15 leads with DAT and SERT}

We are interested in understanding the molecular mechanisms underlying the interactions between lead compounds and critical transporters. To predict the docking poses of these lead compounds with the dopamine transporter (DAT) and serotonin transporter (SERT), we employed the AutoDock Vina docking software \cite{trott2010autodock}. The resulting docking poses are visually represented in Figures \ref{fig:DAT docking} and \ref{fig:SERT docking} for DAT and SERT, respectively.

Moreover, we utilized the LigPlot$+$ software to generate 2D ligand-protein interaction diagrams, illustrating the interactions between these compounds and the two transporters. These interaction diagrams are depicted in Figures \ref{fig:DAT Ligplot} and \ref{fig:SERT Ligplot} for DAT and SERT, respectively. 

How potential drug candidates mediate anti-cocaine addiction can be influenced by their binding sites on the transporters. Notably, the binding site of cocaine on DAT is situated near the center of DAT protein structure \cite{wang2015neurotransmitter}. In the context of treating cocaine addiction, atypical DAT inhibitors stabilize an inward-facing conformation of DAT upon binding. This conformational change hinders the binding of cocaine \cite{reith2015behavioral}. Unlike typical inhibitors, atypical inhibitors do not directly compete with cocaine for the binding site; rather, they diminish the binding effect of cocaine on DAT. Figure \ref{fig:DAT docking} illustrates that some of the lead compounds have docking sites positioned close to the center of DAT structure, while others are located towards the upper portion or in regions away from the central DAT site. For example, compounds 1, 2, 3, 4, 11, 12, and 13 are positioned away from the center. Despite this positioning, these compounds could potentially serve as atypical inhibitors. On the other hand, some other lead compounds have binding sites that are in proximity to the cocaine binding site on DAT. These compounds could potentially exert a more pronounced impact on diminishing the binding effect of cocaine on DAT.

\begin{figure}[htbp!]
    \includegraphics[width=1.0\textwidth]{DAT_Docking_SI.pdf}
    \centering
    \caption{3D structure of molecular docking for candidate leads with DAT (PDB ID: 4XPA). }
    \label{fig:DAT docking}
\end{figure}

Hydrogen bonds play a critical role in the interactions between these compounds and DAT. As depicted in Figure \ref{fig:DAT Ligplot}, among the fifteen protein-ligand complexes, ten display at least one hydrogen bond in their molecular interactions. Notably, both lead 2 and lead 7 establish three distinct hydrogen bonds with DAT. It's worth highlighting that lead 7 is also one of the compounds with binding sites positioned near the center of DAT, as demonstrated in Figure \ref{fig:DAT docking}.

Among the three hydrogen bonds that lead 7 forms with DAT, one involves a hydrogen atom located on the piperazine ring of the compound and an oxygen atom on residue Asp575(A) of DAT. Similarly, this same hydrogen atom on the piperazine ring forms an additional hydrogen bond with an oxygen atom on residue Tyr123(A) of DAT. Furthermore, the same oxygen atom on residue Tyr123(A) of DAT contributes yet another hydrogen bond by interacting with the hydrogen atom of the hydroxyl group on lead 7.

\begin{figure}[htbp!]
    \includegraphics[width=1.0\textwidth]{DAT_Ligplot_SI.pdf}
    \centering
    \caption{Molecular interaction of 15 leads with DAT (PDB ID: 4XPA). }
    \label{fig:DAT Ligplot}
\end{figure}

Cocaine induces addiction by disrupting the regular functioning of SERT as well. Similar to atypical inhibitors for DAT, there are also atypical inhibitors acting on SERT that exert anti-addictive effects. Among these, ibogaine stands out as a noncompetitive inhibitor of both DAT and SERT \cite{bulling2012mechanistic}. It achieves this by stabilizing the transporters in an inward-open conformation. Importantly, ibogaine accomplishes this by binding to a unique binding site on SERT and DAT, distinct from the sites where cocaine or other substrates bind \cite{bulling2012mechanistic}.

The docking sites of the 15 lead compounds on SERT are situated close to the central region of the SERT protein. Notably, the binding site of ibogaine is also located at the central site of SERT, as noted in a study by Coleman et al. \cite{coleman2019serotonin}. In contrast to the varied binding sites observed for these compounds on DAT, their relatively consistent positioning on SERT suggests a potentially more effective modulating effect in combating cocaine addiction.

\begin{figure}[htbp!]
    \includegraphics[width=1.0\textwidth]{SERT_Docking_SI.pdf}
    \centering
    \caption{3D structure of molecular docking for candidate leads with SERT (PDB ID: 6DZZ). }
    \label{fig:SERT docking}
\end{figure}

Figure \ref{fig:SERT Ligplot} illustrates the 2D ligand-protein interaction diagrams between the 15 compounds and SERT. Similar to the molecular interactions between these compounds and DAT, hydrogen bonds play a pivotal role. Among the 15 compounds, eleven are found to form at least one hydrogen bond with SERT. Specifically, Leads 2, 5, and 12 each form three hydrogen bonds with SERT, while lead 4 establishes four hydrogen bonds with SERT.

Interestingly, ten of the 15 lead compounds form hydrogen bonds with DAT, while eleven of them show hydrogen bonds with SERT. Notably, eight compounds share the same trait of forming hydrogen bonds with both DAT and SERT. Based on these observations, we anticipate that effective drug candidates can harness hydrogen bonds to exert binding effects on both DAT and SERT.

\begin{figure}[htbp!]
    \includegraphics[width=1.0\textwidth]{SERT_Ligplot_SI.pdf}
    \centering
    \caption{Molecular interaction of 15 leads with SERT (PDB ID: 6DZZ). }
    \label{fig:SERT Ligplot}
\end{figure}

\section{Supplementary Dialogues}

In this work, we have tailored three persona of ChatGPT to fit three roles within the project: 1) idea generation, 2) methodology elucidation, and 3) coding augmentation. It is worth mentioning that we personified ChatGPT in three individual chatbox. Each individual chatbox does not have access to acquire data from other chatbox. 

Personification refers to the process of assigning human-like characteristics or a persona to an AI model. In this project, we have strategically personified ChatGPT to improve its capacity to better assist our anti-cocaine addiction drug discovery initiative. In this project, we have tailored three persona of ChatGPT to fit three roles within the project: 1) idea generation, 2) methodology elucidation, and 3) coding augmentation. It is worth mentioning that we personified ChatGPT in three individual chatbox. Each individual chatbox does not have access to acquire data from other chatbox. 

\subsection{The 1st persona of ChatGPT}
For the role of idea generation, we assigned ChatGPT the 1st persona of a professor with specific expertise in AI-assisted drug discovery, focusing particularly on treating cocaine addiction. This persona was designed to guide Ph.D. students and postdocs on this specific project, offering insightful explanations, suggestions, or expert advice based on extensive knowledge and experience in the field. The dialogues with the 1st persona is saved in the \href{https://chat.openai.com/share/51d03ee6-331d-47d0-bd0b-5d8bd515a9dd}{1stPersona}.

\subsection{The 2nd persona of ChatGPT}
Moreover, we assigned the 2nd persona of ChatGPT the role of a professional research who is well-versed in diffusion models and statistical methodologies. This persona aims to provide clear explanations, insights, or recommendations in LaTex format. This specific persona was chosen as our 1st ChatGPT persona provided an insightful idea which based on the statistical strategies and diffusion models. The dialogues with the 2nd persona is saved in the \href{https://chat.openai.com/share/8f97de4b-f51b-46d7-903a-d75341ff76c3}{2ndPersona}.

\subsection{The 3rd persona of ChatGPT}
We designated the third persona of ChatGPT as a Python coding specialist, with an emphasis on artificial intelligence and figure generation. This persona is tasked with offering lucid explanations, code snippets, and efficiency optimization for our coding tasks. The dialogues with the 3rd persona is saved in the \href{https://chat.openai.com/share/8256c604-a672-40f5-9668-f7f9e39322f2}{3rdPersona}.

\subsection{Other dialogues}
ChatGPT also assisted us in analyzing SMILES strings. Related dialogues can be found in \href{https://chat.openai.com/share/8539b0c7-438b-4318-9c5f-46886130a44f}{Chemist}.


 \bibliographystyle{unsrt}
 \bibliography{refs}